\documentclass[runningheads]{llncs}

\usepackage[mobile]{eccv}

\usepackage{eccvabbrv}

\usepackage{graphicx}
\usepackage{booktabs}

\usepackage[accsupp]{axessibility}  

\usepackage[dvipsnames]{xcolor}

\usepackage{amsmath}
\usepackage{amssymb}
\usepackage{xcolor}
\usepackage{braket}
\usepackage{bm}
\usepackage{bbm}
\usepackage{array}
\usepackage{colortbl}
\usepackage{soul}

\usepackage{wrapfig}

\newcommand{\rot}[1]{\rotatebox[origin=lb]{60}{\smash{#1}}}

\newcolumntype{x}[1]{>{\centering\arraybackslash}p{#1pt}}
\newcolumntype{y}[1]{>{\raggedright\arraybackslash}p{#1pt}}
\newcolumntype{z}[1]{>{\raggedleft\arraybackslash}p{#1pt}}

\definecolor{mygray}{gray}{0.8}

\usepackage[pagebackref,breaklinks,colorlinks,citecolor=eccvblue]{hyperref}

\usepackage{orcidlink}

\begin{document}

\title{Mind the Interference: Retaining Pre-trained Knowledge in Parameter Efficient Continual Learning of Vision-Language Models} 

\titlerunning{DIKI}

\author{
Longxiang Tang$^1$ \and
Zhuotao Tian$^4$ \and
Kai Li$^5$ \and
Chunming He$^1$ \and
Hantao Zhou$^1$ \and
Hengshuang Zhao$^6$ \and
Xiu Li$^1$\thanks{Corresponding author} \and
Jiaya Jia$^{2,3}$}

\authorrunning{L. Tang et al.}

\institute{
$^1$Tsinghua University\qquad
$^2$SmartMore\qquad
$^3$CUHK\\
$^4$HIT(SZ)\qquad
$^5$Meta Reality Labs\qquad
$^6$HKU
}

\maketitle

\begin{abstract}
    This study addresses the Domain-Class Incremental Learning problem, a realistic but challenging continual learning scenario where both the domain distribution and target classes vary across tasks. To handle these diverse tasks, pre-trained Vision-Language Models (VLMs) are introduced for their strong generalizability. However, this incurs a new problem: the knowledge encoded in the pre-trained VLMs may be disturbed when adapting to new tasks, compromising their inherent zero-shot ability. Existing methods tackle it by tuning VLMs with knowledge distillation on extra datasets, which demands heavy computation overhead. 
    To address this problem efficiently, we propose the Distribution-aware Interference-free Knowledge Integration (DIKI) framework, retaining pre-trained knowledge of VLMs from a perspective of avoiding information interference. Specifically, we design a fully residual mechanism to infuse newly learned knowledge into a frozen backbone, while introducing minimal adverse impacts on pre-trained knowledge. Besides, this residual property enables our distribution-aware integration calibration scheme, explicitly controlling the information implantation process for test data from unseen distributions. Experiments demonstrate that our DIKI surpasses the current state-of-the-art approach using only 0.86\% of the trained parameters and requiring substantially less training time.
    Code is available at: \url{https://github.com/lloongx/DIKI}.
\end{abstract}

\begin{figure}[t]
    \centering
    \includegraphics[width=\linewidth]{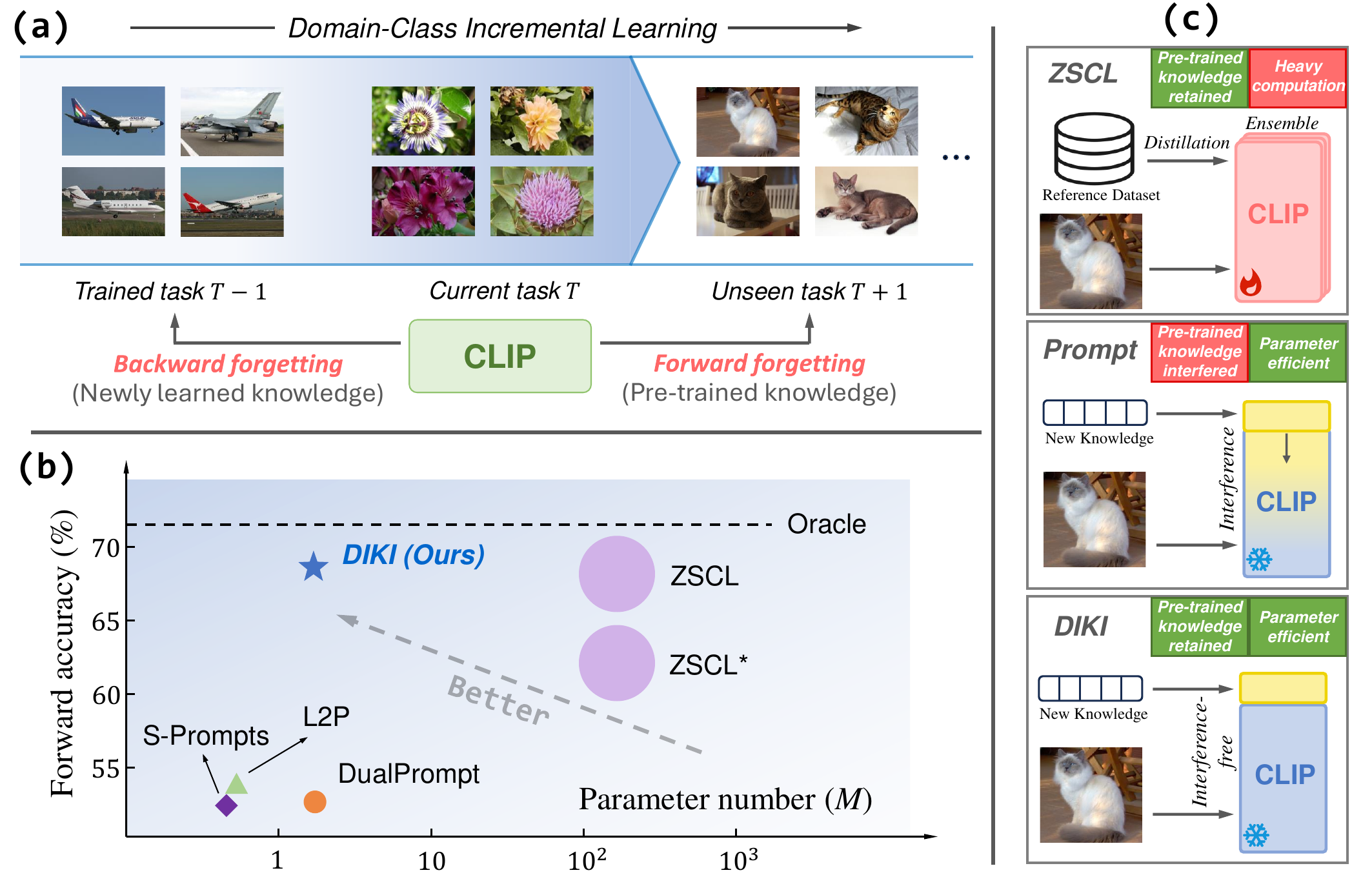} 
    \vspace{-5mm}
    \caption{\textbf{(a)}: The domain-class incremental learning setting, where the data distribution and the classes vary across all tasks. Two kinds of forgetting exist due to the integration of pre-trained CLIP. \textbf{(b)}: The forward accuracy (i.e. zero-shot ability) and the number of trainable parameters for each method, with the size of the markers representing their computational complexity. \textbf{(c)}: Existing methods either demand heavy computation or sacrifice pre-trained knowledge. Our approach effectively retain pre-trained knowledge within a parameter-efficient framework. More details are provided in \cref{sec:IKI}.}
    \label{fig1}
    \vspace{-2mm}
\end{figure}

\section{Introduction}
\label{intro}
Supervised learning techniques train networks with full access to all data, which can result in a lack of flexibility when extending them to acquire knowledge from new tasks. Continual Learning (CL) has emerged as a solution, enabling ongoing model training on sequentially arriving data while retaining the learned information \cite{de2021continual}. Conventional CL settings consider either newly introduced classes or domain distribution shifts, referred to as class incremental and domain incremental learning \cite{van2019three}. However, with only one type of increment considered, these existing works limit their applicability in complex real-world scenarios. 

Consider a more challenging Domain-Class Incremental Learning (DCIL) setting, where both the domain data distribution and classes to be classified can keep varying among all tasks, as illustrated in \cref{fig1}(a). Vanilla image encoder-based techniques are infeasible under such circumstances due to their non-scalable classification head design. Recently, the advent of contrastively trained Vision-Language Models (VLMs), such as CLIP~\cite{radford2021learning}, has made it possible to address this demanding but practical problem. VLMs are trained on web-scale image-text pairs and hold a powerful zero-shot generalization ability to identify nearly infinite classes, making them capable of confronting this severe task variation scenario \cite{zhou2022learning,saharia2022photorealistic,dhariwal2021diffusion,li2022blip,he2024diffusion}.

However, the use of vision-language models introduces new challenges to incremental training. Conventional continual learning schemes aim to prevent models from forgetting previously learned knowledge, which is termed as {\textit{backward forgetting}}~\cite{lopez2017gradient}. Existing works have explored the potential of the regularization mechanism, rehearsal buffer, and architecture design to mitigate backward forgetting, achieving promising results \cite{li2017learning,douillard2022dytox,shin2017continual,rebuffi2017icarl}. Nevertheless, when these approaches are applied to vision-language models, a different form of catastrophic forgetting emerges: \textbf{models tend to forget the knowledge learned during the pre-training phase}, thus compromising their powerful zero-shot generalization capacity. This problem is termed as  \textbf{forward forgetting} because it occurs when VLMs perform ``forward'' prediction on the unknown distributed data. \cref{fig1}(a) illustrates the two types of forgetting.

Recent work ZSCL~\cite{zheng2023preventing} made an attempt to address the forward forgetting issue on CLIP. They introduced a large-scale reference dataset~\cite{deng2009imagenet} to perform knowledge distillation and incorporated a weight ensemble scheme~\cite{wortsman2022robust}. However, this approach requires intensive computation and external data, which could be infeasible in real-world scenarios. Meanwhile, existing VLM-based parameter-efficient continual learning methods \cite{wang2022learning,wang2022dualprompt,wang2022s}, mostly utilizing prompt tuning mechanisms, fail to retain the pre-trained knowledge and cause zero-shot ability degradation, as shown in \cref{fig1}(b).
We attribute this issue to \textbf{information interference}: newly introduced task-specific parameters can disturb the pre-trained knowledge. Illustrations of these methods are shown in \cref{fig1}(c).

To alleviate the forward forgetting problem of VLMs with a computationally and parameter-efficient approach, we introduce the \textbf{Distribution-aware Interference-free Knowledge Integration (DIKI)} framework. Specifically, we inject task-specific information into frozen VLM for each task, storing learned knowledge efficiently.
\underline{(1)} To maintain the pre-trained knowledge in VLMs, our knowledge integration mechanism is designed to resolve the information interference issue prevalent in existing methods. By employing our fully residual design and zero-initialization strategy, we can inject new knowledge while keeping the pre-trained knowledge untouched, introducing minimal noise to the pre-trained model compared to prompt tuning. \underline{(2)} With this advantage, we further introduce a distribution-aware integration calibration mechanism, explicitly identifying the unseen distributed samples and controlling the implanted information for them, thereby enhancing the model generalization capabilities.

Our contributions are summarized in threefold:
\begin{itemize}
    \item We introduce the parameter-efficient DIKI to retain pre-trained knowledge in VLMs under the DCIL setting. It resolves the information interference issue, mitigating the need for heavy computation and external data.
    \item To alleviate the forward forgetting, DIKI implants new knowledge in a fully residual manner, leaving pre-trained knowledge undisturbed. With this residual property, a distribution-aware integration calibration is incorporated to further boost performance on unseen tasks.
    \item Comprehensive experiments demonstrate that we achieve state-of-the-art performance with only 0.86\% trained parameters and significantly less training time compared to the previous methods.
\end{itemize}


\section{Related Works}

\noindent\textbf{Continual learning.}
Existing continual learning algorithms can be broadly classified into three categories~\cite{de2021continual}. \textit{Regularization-based} methods \cite{li2017learning,zhang2020class,kirkpatrick2017overcoming,ahn2019uncertainty,aljundi2018memory,lai2021semi} introduce an extra regularization term in the loss function, consolidating previous knowledge when learning on new data. In contrast, \textit{architecture-based} methods \cite{douillard2022dytox,yoon2017lifelong,li2019learn,rao2019continual,mallya2018packnet} dedicate different model parameters to each task, storing task knowledge with specific expanded network components. With memory replay technique, \textit{rehearsal-based} methods \cite{shin2017continual,lopez2017gradient,rebuffi2017icarl,isele2018selective,rolnick2019experience} retrain current step model with stored exemplars in raw format or generated pseudo-samples with a generative model, which has been questioned for its rationality by recent work~\cite{prabhu2020gdumb}. While achieving promising results, these solutions only consider one type of increment, either domain shift or new classes, along the continual training process, resulting in limited applicability in real-world scenarios. Instead, we investigate the forgetting problem under a domain-class incremental learning setting to adapt to a broader variety of situations.

\noindent\textbf{Parameter-efficient fine-tuning.}
Fully fine-tuning a large pre-trained model is computationally expensive and requires a large-scale dataset~\cite{wortsman2022robust}. Alternatively, parameter-efficient fine-tuning approaches only introduce a small set of parameters to rapidly adapt a pre-trained model to downstream tasks, such as LoRA \cite{hu2021lora}, prompt tuning \cite{jia2022visual,liu2023pre,sohn2023visual,yang2024exploring} and adapters \cite{wang2020kadapter,houlsby2019parameter}. Due to their simple and portable design, prompt tuning techniques have attracted many applications in a variety of areas~\cite{gao2023clip,ju2022prompting,zhou2022learning,khattak2023maple,hegde2023clip}. However, existing prompt learning-based methods typically prepend the learnable parameters to the original input tokens, where lies the information interference issue and eventually causes pre-trained knowledge loss during the training process.

\noindent\textbf{Vision-language models.}
Trivial visual-only models extract features from images and then utilize a fixed head to derive final predictions, constraining their flexibility across tasks \cite{he2016deep,he2023strategic,fang2024real,he2023camouflaged,he2023reti,tian2022generalized}. Vision-Language Models (VLMs) present a solution by leveraging the interaction between image and text descriptions \cite{radford2021learning,jia2021scaling,zhai2022lit,yao2021filip,zhou2024unihead,lai2024lisa,yang2024v,yang2024unified}. Trained on web-scale image-text pair datasets, V-L models can identify nearly infinite classes and can be easily transferred to unseen domains, holding a strong zero-shot ability. However, most previous VLMs continual learning methods \cite{wang2022s,smith2023construct,khan2023introducing,qian2023decouple,zhou2023learning} have not considered the zero-shot performance drop during the training process, which can cause a significant model degradation towards unseen data distributions.

\section{Preliminaries}

\textbf{Continual learning protocol.}
Continual learning aims to sequentially learn different tasks without forgetting previously learned knowledge.
Considering $N$ sequentially arrived tasks $\left[ \mathcal{T}^1, \mathcal{T}^2, \cdots, \mathcal{T}^N \right]$, each task $\mathcal{T}^i$ contains a dataset $D^i=\{x^i_j, y^i_j\}_{j=1}^{N^i}$, where $x^i_j$ is an image and $y^i_j$ is corresponding one-hot label inside current dataset, and $N_i$ is the number of image samples. Additionally, a class name set $C^i=\{c^i_j\}_{j=1}^{N_{c}^i}$ is included, linking the label index to a category name used by the VLMs.

Different from previous class- and domain-incremental learning settings, this work highlights a more practical continual learning setting: Domain-Class Incremental Learning (DCIL). In this setting, domain distribution and classes to be identified keep varying among different tasks, i.e. $C^i \neq C^j$ and $\mathbb{P}(D^i) \neq \mathbb{P}(D^j)$ for $i \neq j$, where $\mathbb{P}$ represents data distribution of a task dataset.

\noindent\textbf{Vision-language models.} Towards the challenging DCIL setting, training a vanilla image encoder-based model, such as ResNets \cite{he2016deep} and ViTs \cite{dosovitskiy2020image}, is not practical for incrementally learning intensely shifted domains and classes. Hence, pre-trained vision-language models are introduced for their robust zero-shot transfer capabilities. 
CLIP~\cite{radford2021learning} consists of an image encoder $f$ and a text encoder $g$, which are trained to generate closely aligned feature representations for paired image-text samples. At inference time, $f$ first encodes the input image $x$ into a feature vector $f(x)$. Concurrently, potential class names are embedded into a template, like ``a photo of \{$c$\}'', and then encoded by $g$ to form text embeddings $\{t_j\}_{j=1}^{N_c}$. The model predictions are determined by the largest similarity scores between image embedding and all text embeddings, formulated as $s_j = \Braket{f(x), t_j}$, where $\Braket{\cdot, \cdot}$ denotes the cosine similarity.

\noindent\textbf{Task-specific prompt learning.}
Following the success of \cite{wang2022learning,wang2022dualprompt}, a series of works \cite{smith2023coda,hu2023pop,bowman2023carte} begin to explore the potential of parameter-efficient fine-tuning in continual learning. A common practice is learning and storing a set of lightweight prompts for each task, forming a ``prompt pool'' during the continual learning phase, formulated as:
\begin{equation}
    \mathbf{P}=\{P_1, P_2, \cdots, P_N\},\ \ \text{where}\ P_i\in \mathbb{R}^{l\times d},
\end{equation}
where $N$ is the task number, $l$ and $d$ are the prompt length and the feature embedding dimension.

At inference time, well-trained prompts are selected and attached to the pre-trained frozen model, restoring the learned knowledge. Assume $\bm{x_e}\in \mathbb{R}^{L\times d}$ is the feature embeddings for a transformer layer $h$, then we can prepend the prompts to the $\bm{x_e}$ to generate prompted inputs:
\begin{equation}
    \bm{x_p} = \left[P_s^1; P_s^2; \cdots; P_s^l; \bm{x_e}\right] \in \mathbb{R}^{(l+L)\times d},
\end{equation}
where $\{P_s^i\in \mathbb{R}^{d}\}_{i=1}^l$ are embedding vectors of selected prompt $P_s$ and $;$ represents the concatenation operation along the token length dimension. With this implanted knowledge, better image and text feature embeddings are generated, and the final classification accuracy is improved.

The prompt selection process mentioned above is implemented by query-key matching. During the continual training stage, average feature representations $\mathbf{I}=\{I^i\}_{i=1}^N$ for each task are learned by maximizing cosine similarity~\cite{wang2022learning,wang2022dualprompt} or by applying clustering algorithm~\cite{wang2022s}. When a test sample $\bm{x}$ comes, a key lookup regime is performed:
\begin{equation}
\label{eq_matching}
    I_s = {\arg \max}_{I^i\sim \mathbf{I}}\Braket{f(\bm{x}), I^i}.
\end{equation}

With the most relevant key $I_s$, corresponding prompts $P_s$ are selected and attached to the frozen model, performing inference process.

\section{Methodology}

\subsection{Interference-free Knowledge Integration}
\label{sec:IKI}
\textbf{Is prepending the best choice?}
Despite methods that prepend prompt to input tokens are widely used for their simplicity in implementation, we identified that they are suffering from issues in two folds.

Firstly, concatenating the prompts with input tokens causes them to interact during the attention process, and influences the pre-trained knowledge extraction, which will be discussed below. 
When the test samples are drawn from the distribution where the model learned the prompts, the adapted model can preserve relatively satisfactory results. However, once encountering samples with a distribution shift, this interference could result in model degradation and a loss of its vital zero-shot generalization ability, causing forward forgetting issues.

Besides, simply prepending prompts inevitably increases the token length across all transformer blocks, which is not desirable in many scenarios with token length constraints. In addition, its scalability is limited: a long prompt context can distract the text encoder from informative class names, resulting in poor text embedding representation.

The existence of the above issues indicates that prompt tuning-based methods do not satisfy the ``residual property'': we expect learned parameters should be a residual path paralleled to the frozen backbone, supplementing novel knowledge without affecting the crucial pre-trained knowledge. 
Therefore, we propose a \textit{Interference-free Knowledge Integration (IKI)} scheme to inject newly learned knowledge into a pre-trained VLM with introducing minimal noise to it.

\noindent\textbf{IKI mechanism.} Instead of training a series of prepended prompt vectors for each task, we focus on self-attention mechanism modification following widely used parameter efficient fine-tuning methods in NLP field~\cite{hu2021lora,li2021prefix,zhang2023llama,tian2019learning}.
Recall the multi-head self-attention~\cite{vaswani2017attention} mechanism conducted on input tokens $\bm{x_e}\in \mathbb{R}^{L\times d}$ in transformer layer $h$. For simplification, we omit the multi-head design and solely consider the one-head situation, which can be naturally extended to multi-head scenarios. Input tokens are first transformed to query $Q$, key $K$ and value $V$ matrices by linear projections:
\begin{equation}
    Q_e = \bm{x_e}W^Q + b^Q; K_e = \bm{x_e}W^K + b^K; V_e = \bm{x_e}W^V + b^V,
\end{equation}
where $W\in \mathbb{R}^{d\times d}$ and $b\in \mathbb{R}^{d}$ are pre-trained parameters. Then self-attention calculation is performed to produce an output matrix via
\begin{equation}
    O_L = \text{Attn}(Q_e, K_e)V_e = \text{softmax}(\frac{Q_eK_e^T}{\sqrt{d}})V_e\ \ \in \mathbb{R}^{L\times d},
\end{equation}
where $\text{softmax}(\bm{z})_i = \frac{\exp{(\bm{z_i})}}{\sum_j\exp{(\bm{z_j})}}$ can constrain the elements in attention results $\text{Attn}(Q_e, K_e)\in \mathbb{R}^{L\times L}$ sum to one.

\begin{figure*}[t]
    \centering
    \includegraphics[width=\linewidth]{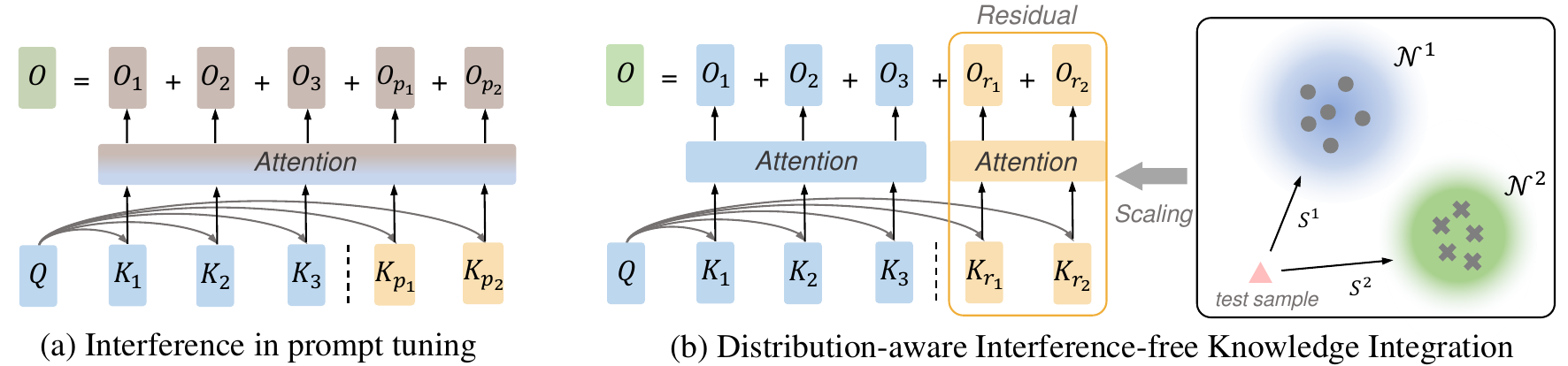} 
    \caption{Illustration of the information interference issue in previous prompt tuning methods and our proposed DIKI. (a) The existing methods mix attention derived from the frozen backbone and prepended prompts, which can cause information loss and finally harm the zero-shot ability. (b) We design a zero-initialized residual attention mechanism, which injects new information with pre-trained knowledge untouched, to retain the vision-language models' zero-shot ability. Distribution-aware integration calibration is also introduced to further boost performance thanks to the residual property.}
    \label{fig:framework}
\end{figure*}

Vanilla prompt tuning methods prepend trainable prompts to input tokens, extending $\bm{x_e}\in \mathbb{R}^{L\times d}$ to $\bm{x_p}\in \mathbb{R}^{(l+L)\times d}$. Then $Q_{p}K_{p}^T\in \mathbb{R}^{(l+L)\times (l+L)}$ will be computed and passed to a softmax function. Inside softmax calculation, attention scores of input tokens and prompts interact and affect each other, leading to an inevitable loss of pre-trained knowledge, as illustrated in \cref{fig:framework} (a).

To address this issue, we compute attention outputs for self-attention within input tokens and cross-attention between prompts and input tokens separately, as shown in \cref{fig:framework} (b). In other words, we only train a residual attention branch, leaving the existing attention score untouched. With newly introduced keys $K_r$ and values $V_r$, the output of our residual attention branch can be formulated as:
\begin{equation}
    \label{eq:res_attn}
    O_r = \text{softmax}(\frac{Q_eK_r^T}{\sqrt{d}})V_r, \text{where}\ K_r,V_r\in \mathbb{R}^{l\times d}.
\end{equation}

Here the residual output $O_r\in \mathbb{R}^{L\times d}$ is derived with an orthogonal path to the original output $O_L$, producing no influence on the original attention process. Finally, the learned knowledge stored in $O_r$ is implanted into output by addition. During continual training stage, we update the learnable keys $K_r$ and values $V_r$ instead of commonly used prompts $P$. Note that to keep sequence length unchanged, we didn't introduce any query parameters.

Ideally, a desirable residual block should not affect the original branch before being trained on downstream datasets, i.e. at initialization time. Widely used protocols initialize prompts with uniform or normal distribution, which injects random noise into the pre-trained VLMs even when no knowledge has been learned. Specifically, we enforce residual attention addition to be an identity function by zero-initialize the parameters $V_r$:
\begin{equation}
    O = O_L+O_r^{\text{init}} = O_L+\text{softmax}(\frac{Q_eK_r^T}{\sqrt{d}})\mathbf{[0]}^{l\times d} = O_L.
\end{equation}

Note that we only constrain values $V_r^{\text{init}}$ to be zero at the beginning, while keeping $K_r$ random initialized. That's because initializing both $K_r$ and $V_r$ to zero matrix will prevent $K_r$ from updating by gradient flow, and make $V_r$ degenerate to vectors with same values. We prove this in the supplementary materials.

Since zero-initialization is more like a choice rather than a technique, some studies \cite{chen2022adaptformer,jie2023fact,zhang2023llama} have adopted it across various tasks. However, these works leverage it to ensure a stable and progressive training regime, a concern that is not present in DCIL scenarios. We argue that zero-initialization is essential for our residual attention design to inject new knowledge into the pre-trained VLMs with minimal noise introduced, which is demonstrated in \cref{sec:analysis}.


\subsection{Distribution-aware Integration Calibration}

\noindent\textbf{Observations.}
At inference time, the query-key matching mechanism described in \cref{eq_matching} is performed to retrieve appropriate learned prompts for the current test sample. This approach is tailored for conventional continual learning settings, which only considers the backward forgetting mentioned in \cref{intro}. However, when confronted with data from unseen domains, this trivial matching design is enforced to assign a relatively similar task for test samples, despite there's a significant distribution gap between them.

Benefiting from the residual design of our proposed IKI, we can introduce less noise in such mismatch scenarios compared with previous methods. Nonetheless, when the discrepancy between training and testing distribution increases, it's inevitable to cause model degradation to some extent and hurt the zero-shot ability that VLMs learned during the pre-train phase.

ZSCL~\cite{zheng2023preventing} tackles this problem via distillation. They build a reference dataset with \textit{100k} images from ImageNet \cite{deng2009imagenet} to distill pre-trained knowledge from the original CLIP to the current model at every training step, explicitly performing rehearsal to avoid forgetting. This approach could be effective, but it relies on large-scale storage and high computation resources as shown in \cref{tab:computation_comp}, making it impractical under real-world circumstances.

One intuitive solution to this issue is controlling to what extent knowledge is implanted into the model. However, previous prepending-based prompt tuning techniques have only two choices: either appending learned prompts or leaving the original CLIP model untouched. Thanks to the graceful residual property from our IKI, we obtain the ability to control this paralleled branch.

\noindent\textbf{DIKI: calibrate the integration with distribution.}
To determine the likelihood that a test sample belongs to a learned task, we maintain a feature distribution \cite{tang2023consistency,tang2023source,he2023weaklysupervised,tian2020prior,peng2023hierarchical} instead of a single key vector for every task. Here we simply apply multivariate Gaussian distribution and find it works well. Formally, we build a $\mathcal{N}^i(\bm{\mu}^i, \bm{\Sigma}^i)$ for task $i$ during training stage:
\begin{equation}
\begin{gathered}
    \bm{\mu}^i = \mathbb{E}_{\bm{x}^i_j \sim D^i}[f(\bm{x}^i_j)], \ \ \ \bm{\Sigma}^i = \mathbb{E}_{\bm{x}^i_j \sim D^i}[(f(\bm{x}^i_j)-\bm{\mu}^i)^T(f(\bm{x}^i_j)-\bm{\mu}^i)],
\end{gathered}
\end{equation}
where $f(\bm{x}^i_j)$ is the image feature extracted by frozen encoder. With these estimated distributions, the possibility of a test sample being drawn from each $\mathcal{N}^i$ can be calculated. Here we compute the logarithm of the probability density as a scoring function for input $\bm{x}$ on each learned task:
\begin{equation}
\begin{split}
    S^i &= \log \varphi(f(\bm{x}); \bm{\mu}^i, \bm{\Sigma}^i) \\
    &= - \frac{1}{2}[ (f(\bm{x})-\bm{\mu}^i)^T(\bm{\Sigma}^i)^{-1}(f(\bm{x})-\bm{\mu}^i) + d\log 2\pi + \log |\bm{\Sigma}^i|) ],
\end{split}
\end{equation}
where $\varphi$ is the probability density function.

Intuitively, a sample with a higher score $S^i$ is more likely to be drawn from task $i$, and parameters $K_r^i, V_r^i$ should be introduced for model prediction. Besides, we should also take into account that income sample $\bm{x}$ might come from some new distributions, which is suggested if all $S^i$ are low. Thus we utilize the maximum score $\hat{S}=\max_{i\in [1,N]}S^{i}$ to weight the residual attention output:
\begin{equation}
    \label{eq:final_output}
    O = O_L+\mathcal{M}(\hat{S})O_r,
\end{equation}
where $\mathcal{M}$ is a mapping function that scales the score $\hat{S}$ to the range $[0,1]$. Here we find a simple Sigmoid function $\sigma(x)=\frac{1}{1+e^{-x}}$ works well here. We also conduct experiments in \cref{sec:analysis} to demonstrate the rationality and correctness of the calibration technique on IKI outputs.

Empowered by this distribution-aware integration calibration mechanism, the pre-trained zero-shot ability of VLMs can be retained better by assign lower weight to unfamiliar images, further resolving the forward forgetting issue.

\section{Experiments}

\textbf{Benchmarks.} To demonstrate the effectiveness of DIKI under the domain-class incremental learning setting, we conduct experiments on the recently proposed MTIL~\cite{zheng2023preventing} benchmark. MTIL consists of 11 diverse datasets: Aircraft \cite{maji2013fine}, Caltech101 \cite{fei2004learning}, CIFAR100 \cite{krizhevsky2009learning}, DTD \cite{cimpoi2014describing}, EuroSAT \cite{helber2019eurosat}, Flowers \cite{nilsback2008automated}, Food \cite{bossard2014food}, MNIST \cite{deng2012mnist}, OxfordPet \cite{parkhi2012cats}, StanfordCars \cite{krause20133d}, and SUN397 \cite{xiao2010sun}. It's a very challenging benchmark with total of 1201 classes and severe data distribution shift across different tasks, which is infeasible for vanilla image encoder-based methods. Thus, vision-language models are necessarily included. The Order-I in original paper is applied. We also introduce the modified MTIL-FS benchmark for few-shot setting evaluation, in which only 16 samples per class of each dataset are used for training to simulate the data deficient scenario. More details can be found in the supplementary materials.

\begin{table*}[t]
\setlength\tabcolsep{5pt}
\centering
\setlength{\belowcaptionskip}{2mm}
\caption{\textit{Transfer}, \textit{Avg.}, and \textit{Last} scores (\%) of different continue learning methods on MTIL benchmark. Metric ``transfer'' represents the model zero-shot ability retention after being trained on each task. $\dag$ means we reproduce the original methods on vision-language models. 
}
\label{tab:finalacc}
{
\fontsize{8pt}{10pt}\selectfont
\resizebox{1\textwidth}{!}{
\begin{tabular}{y{70}x{8}x{25}|*{11}{x{17}}|x{22}}
\toprule
 & \rot{Extra data} & \rot{\# Param.} & \rot{Aircraft} & \rot{Caltech101} & \rot{CIFAR100} & \rot{DTD} & \rot{EuroSAT} & \rot{Flowers} & \rot{Food} & \rot{MNIST} & \rot{OxfordPet} & \rot{Cars} & \rot{SUN397} & \rot{Average} \\ \midrule

\quad Zero-shot & & & 24.8 & 92.9 & 68.4 & 43.8 & 47.7 & 71.4 & 85.8 & 59.5 & 89.1 & 65.8 & 62.6 & 64.7 \\
\quad Upper Bound & & & 62.0 & 96.2 & 89.6 & 79.5 & 98.9 & 97.5 & 92.7 & 99.6 & 94.7 & 89.6 & 81.8 & 89.3 \\ \midrule
\midrule

\textbf{Transfer} \\
\quad LwF~\cite{li2017learning} & $\checkmark$ & 211 M & & 74.5 & 56.9 & 39.1 & \textbf{51.1} & 52.6 & 72.8 & 60.6 & 75.1 & 30.3 & 55.9 & 56.9 \\
\quad iCaRL~\cite{rebuffi2017icarl} & $\checkmark$ & 211 M & & 56.6 & 44.6 & 32.7 & 39.3 & 46.6 & 68.0 & 46.0 & 77.4 & 31.9 & 60.5 & 50.4 \\
\quad LwF-VR~\cite{ding2022don} & $\checkmark$ & 211 M & & 77.1 & 61.0 & 40.5 & 45.3 & 54.4 & 74.6 & 47.9 & 76.7 & 36.3 & 58.6 & 57.2 \\
\quad WiSE-FT~\cite{wortsman2022robust} & $\checkmark$ & 211 M & & 73.5 & 55.6 & 35.6 & 41.5 & 47.0 & 68.3 & 53.9 & 69.3 & 26.8 & 51.9 & 52.3  \\
\quad ZSCL$^*$~\cite{zheng2023preventing} & $\checkmark$ & 211 M & & 78.3 & 64.0 & 42.9 & 45.2 & 63.5 & 84.2 & 56.1 & 78.9 & 44.1 & 64.3 & 62.2 \\
\quad ZSCL~\cite{zheng2023preventing} & $\checkmark$ & 211 M & & 86.0 & 67.4 & \textbf{45.4} & 50.4 & \textbf{69.1} & \textbf{87.6} & 61.8 & 86.8 & 60.1 & \textbf{66.8} & 68.1 \\ \midrule
\quad L2P$^\dag$~\cite{wang2022learning} & $\times$ & 0.5 M & & 65.6 & 50.9 & 30.4 & 41.4 & 49.3 & 71.8 & 36.3 & 77.5 & 55.3 & 53.4 & 53.2 \\
\quad DualPmt.$^\dag$\hspace{-1mm}~\cite{wang2022dualprompt} & $\times$ & 1.8 M & & 56.7 & 51.4 & 28.7 & 33.7 & 45.6 & 70.9 & 59.5 & 77.7 & 49.5 & 50.4 & 52.4 \\
\quad S-Prompts~\cite{wang2022s} & $\times$ & 0.5 M & & 67.3 & 49.4 & 26.4 & 39.7 & 47.1 & 70.2 & 34.3 & 78.9 & 56.7 & 52.2 & 52.2 \\
\rowcolor{mygray} \quad DIKI & $\times$ & 1.8 M & & \textbf{92.9} & \textbf{69.0} & 43.2 & 48.2 & 67.4 & 85.2 & \textbf{63.0} & \textbf{87.9} & \textbf{63.8} & 66.2 & \textbf{68.7} \\ \midrule
\midrule

\textbf{Avg.} \\
\quad LwF~\cite{li2017learning} & $\checkmark$ & 211 M & 36.3 & 86.9 & 72.0 & 59.0 & 73.7 & 60.0 & 73.6 & 74.8 & 80.0 & 37.3 & 58.1 & 64.7 \\
\quad iCaRL~\cite{rebuffi2017icarl} & $\checkmark$ & 211 M & 35.5 & 89.2 & 72.2 & 60.6 & 68.8 & 70.0 & 78.2 & 62.3 & 81.8 & 41.2 & 62.5 & 65.7 \\
\quad LwF-VR~\cite{ding2022don} & $\checkmark$ & 211 M & 29.6 & 87.7 & 74.4 & 59.5 & 72.4 & 63.6 & 77.0 & 66.7 & 81.2 & 43.7 & 60.7 & 65.1 \\
\quad WiSE-FT~\cite{wortsman2022robust} & $\checkmark$ & 211 M & 26.7 & 86.5 & 64.3 & 57.1 & 65.7 & 58.7 & 71.1 & 70.5 & 75.8 & 36.9 & 54.6 & 60.7 \\
\quad ZSCL$^*$~\cite{zheng2023preventing} & $\checkmark$ & 211 M & \textbf{50.7} & 90.9 & 79.8 & 63.8 & 76.6 & 77.3 & 87.0 & 71.9 & 83.0 & 52.0 & 65.9 & 72.6 \\
\quad ZSCL~\cite{zheng2023preventing} & $\checkmark$ & 211 M & 45.1 & 92.0 & 80.1 & 64.3 & 79.5 & 81.6 & \textbf{89.6} & 75.2 & 88.9 & 64.7 & \textbf{68.0} & 75.4 \\ \midrule
\quad L2P$^\dag$~\cite{wang2022learning} & $\times$ & 0.5 M & 38.0 & 85.2 & 78.2 & 61.3 & 72.9 & 74.9 & 79.7 & 59.1 & 82.0 & 59.7 & 55.4 & 67.9 \\
\quad DualPmt.$^\dag$\hspace{-1mm}~\cite{wang2022dualprompt} & $\times$ & 1.8 M & 37.8 & 84.3 & 78.6 & 60.1 & 71.1 & 73.2 & 79.1 & 73.9 & 82.3 & 55.1 & 52.8 & 68.0 \\
\quad S-Prompts~\cite{wang2022s} & $\times$ & 0.5 M & 37.5 & 92.5 & 77.5 & 58.2 & 76.4 & 74.1 & 78.8 & 57.9 & 83.0 & 60.8 & 54.4 & 68.3 \\
\rowcolor{mygray} \quad DIKI & $\times$ & 1.8 M & 45.1 & \textbf{95.5} & \textbf{83.1} & \textbf{64.8} & \textbf{79.9} & \textbf{83.5} & 87.0 & \textbf{76.2} & \textbf{89.6} & \textbf{67.0} & 67.1 & \textbf{76.3} \\ \midrule
\midrule

\textbf{Last} \\
\quad LwF~\cite{li2017learning} & $\checkmark$ & 211 M & 26.3 & 87.5 & 71.9 & 66.6 & 79.9 & 66.9 & 83.8 & \textbf{99.6} & 92.1 & 66.1 & 80.4 & 74.6 \\
\quad iCaRL~\cite{rebuffi2017icarl} & $\checkmark$ & 211 M & 35.8 & 93.0 & 77.0 & 70.2 & 83.3 & 88.5 & 90.4 & 86.7 & 93.2 & 81.2 & \textbf{81.9} & 80.1 \\
\quad LwF-VR~\cite{ding2022don} & $\checkmark$ & 211 M & 20.5 & 89.8 & 72.3 & 67.6 & 85.5 & 73.8 & 85.7 & \textbf{99.6} & 93.1 & 73.3 & 80.9 & 76.6 \\
\quad WiSE-FT~\cite{wortsman2022robust} & $\checkmark$ & 211 M & 27.2 & 90.8 & 68.0 & 68.9 & 86.9 & 74.0 & 87.6 & \textbf{99.6} & 92.6 & 77.8 & 81.3 & 77.7 \\
\quad ZSCL$^*$~\cite{zheng2023preventing} & $\checkmark$ & 211 M & \textbf{46.0} & 92.3 & 81.2 & 72.4 & 93.0 & 92.1 & 90.8 & \textbf{99.6} & 93.3 & \textbf{86.6} & 81.7 & 84.5 \\
\quad ZSCL~\cite{zheng2023preventing} & $\checkmark$ & 211 M & 40.6 & 92.2 & 81.3 & 70.5 & 94.8 & 90.5 & \textbf{91.9} & 98.7 & 93.9 & 85.3 & 80.2 & 83.6 \\
\midrule
\quad L2P$^\dag$~\cite{wang2022learning} & $\times$ & 0.5 M & 38.0 & 87.1 & 84.2 & 72.9 & 86.0 & 96.1 & 89.2 & 99.0 & 94.1 & 79.6 & 76.0 & 82.0 \\
\quad DualPmt.$^\dag$\hspace{-1mm}~\cite{wang2022dualprompt} & $\times$ & 1.8 M & 37.8 & 87.1 & 84.6 & 71.8 & 89.2 & 96.3 & 89.1 & 99.1 & \textbf{94.5} & 79.9 & 76.5 & 82.3 \\
\quad S-Prompts~\cite{wang2022s} & $\times$ & 0.5 M & 37.5 & 95.1 & 83.7 & 70.2 & 97.5 & 96.5 & 89.0 & 99.1 & 94.0 & 79.5 & 75.8 & 83.4 \\
\rowcolor{mygray} \quad DIKI & $\times$ & 1.8 M & 45.2 & \textbf{95.7} & \textbf{86.3} & \textbf{72.9} & \textbf{98.0} & \textbf{97.0} & 89.2 & 99.4 & 94.2 & 81.6 & 76.6 & \textbf{85.1} \\ 
\bottomrule
\end{tabular}}%
}
\vspace{-2mm}
\end{table*}

\noindent\textbf{Evaluation metrics.} To evaluate both backward and forward forgetting issues mentioned in \cref{intro}, we adopt \textit{Transfer}, \textit{Avg.} and \textit{Last} metrics from \cite{zheng2023preventing}. \textit{Last} score is the model performance after all continual training, representing the degree of backward forgetting and being widely used in conventional continual learning. For forward forgetting issues, i.e. the loss of zero-shot ability, we evaluate model average accuracy on task $i+1, i+2, ..., N$ after its training on task $i$, denoted by \textit{Transfer}. Lastly, \textit{Avg.} is the average accuracy across all time steps. Detailed formulations can be found in the supplementary materials.

\noindent\textbf{Comparison methods.} We compare our DIKI against both full-parameter fine-tuning and parameter-efficient fine-tuning methods. For full fine-tuning, we choose ZSCL, ZSCL*~\cite{zheng2023preventing}, LwF~\cite{li2017learning}, iCaRL~\cite{rebuffi2017icarl}, LwF-VR~\cite{ding2022don}, and WiSE-FT~\cite{wortsman2022robust} following \cite{zheng2023preventing}. For parameter efficient ones, L2P~\cite{wang2022learning}, DualPrompt~\cite{wang2022dualprompt}, and S-Prompts~\cite{wang2022s} are selected for the similar task-specific parameter training procedure to our DIKI. Note that original L2P and DualPrompt are designed for ViT~\cite{dosovitskiy2020image}, we reproduce them on CLIP for fair comparisons. More reproduction details can be found in the supplementary materials.

\noindent\textbf{Implementation details.} We adopt CLIP ViT-B/16 \cite{radford2021learning} as our vision-language model for fair comparisons. In the training process, we optimize the cross entropy loss between model prediction and ground truth. SGD optimizer with cosine learning rate scheduler is applied for all experiments, and the learning rate and batch size are set to 5 and 128, separately. Models are trained with 10 epochs on each task. For trainable parameters $K_r$ and $V_r$, we set both the length $l$ and training layer depth to 8 as discussed in the supplementary materials. To avoid floating point arithmetic precision problems, a small number $10^{-7}$ is added to diagonal elements of covariance matrix $\Sigma^i$ with minor influence on final accuracy. All experiments are conducted on one NVIDIA 3090 GPU.

\begin{table}[t]

\begin{minipage}[!t]{0.49\textwidth}
    \centering
    \setlength{\belowcaptionskip}{2mm}
    \caption{Transfer, Avg., and Last scores (\%) of different continual learning methods on 16-shot MTIL-FS benchmark. Full results can be found in the supplementary materials. Our DIKI can achieve more improvement when data is insufficient due to its non-interfered knowledge implantation scheme. $\dag$ is equivalent to \cref{tab:finalacc}.
    }
    \label{tab:fewshot}
    \resizebox{\textwidth}{!}{%
    {
        \fontsize{8pt}{10pt}\selectfont
        \begin{tabular}{y{70}|x{26}x{26}x{26}|x{30}}
        \toprule
        & Trans. & Avg. & Last & Average \\
        \midrule
        Zero-shot & 70.1 & - & - & - \\
        \midrule
        ZSCL~\cite{zheng2023preventing} & 68.3 & 69.3 & 74.0 & 70.5 \\
        \midrule
        L2P$^\dag$~\cite{wang2022learning} & 53.9 & 62.3 & 73.3 & 63.2 \\
        DualPrompt$^\dag$~\cite{wang2022dualprompt} & 57.9 & 64.3 & 74.7 & 65.6 \\
        S-Prompts~\cite{wang2022s} & 55.5 & 63.2 & 73.8 & 64.2 \\
        \rowcolor{mygray} DIKI & \textbf{70.3} & \textbf{71.9} & \textbf{77.1} & \textbf{73.1} \\
        \bottomrule
        \end{tabular}
    }
    }
\end{minipage}%
\hspace{0.02\textwidth}
\begin{minipage}[!t]{0.49\textwidth}
    \centering
    \setlength{\belowcaptionskip}{2mm}
    \caption{Ablation study of DIKI's components on MTIL benchmark. Our proposed modules form an integrated whole: zero-initialization only works with our residual attention design, and the calibration technique is designed on top of the residual branch. Note that our zero-initialization and calibration techniques only affect zero-shot ability, i.e. \textit{Transfer} metric.
    }
    {
    \fontsize{8pt}{10pt}\selectfont
    \resizebox{\textwidth}{!}{
        \begin{tabular}{x{30}x{34}x{27}x{27}|x{26}x{26}}
            \toprule
            Prompt &  ResAttn & Z-init & Calib. & Transfer & Last \\
            
            \midrule
            
            $\checkmark$& & & & 57.7 & 84.1 \\
            $\checkmark$ & & $\checkmark$ & & 57.3 & 84.0 \\
             & $\checkmark$ &  & & 59.9 & 85.2 \\
             & $\checkmark$ & $\checkmark$ & & 63.1 & 85.0 \\
            \rowcolor{mygray}  & $\checkmark$ & $\checkmark$ & $\checkmark$ & 68.7 & 85.1 \\
            
            \bottomrule
        \end{tabular}
    }
    }
    \label{tab:main_ablation}
\end{minipage}

\vspace{-2mm}

\end{table}

\subsection{Main Results}

\cref{tab:finalacc} contains the \textit{Transfer}, \textit{Avg.} and \textit{Last} scores among all methods on MTIL benchmark. ``Extra data'' includes memory buffers and reference datasets which used in distillation \cite{zheng2023preventing}, and ``\# Param.'' is the number of trainable parameters. ``Zero-shot'' results are simply derived from leveraging the original CLIP weight on each task and perform as a comparison reference for \textit{Transfer} metric. Note that \textit{Transfer} scores can be higher than zero-shot results, because knowledge learned from current task $i$ may contain some task-invariant information which can boost the performance of future tasks $i+1, i+2, ..., N$. ``Upper Bound'' is calculated by applying full parameter fine-tuning technique on each separate dataset, as a guide for \textit{Last} score. 

As indicated by the bold values, our DIKI outperforms the previous state-of-the-art method \cite{zheng2023preventing} across all three metrics with only 0.86\% trainable parameters, while alleviating the requirement for any external data. Thanks to the task-specific parameter training technique, we can memorize previous tasks' knowledge without rehearsal buffers and parameter ensemble, maintaining a high \textit{Last} score with low computational complexity. Moreover, compared with task-specific prompt tuning methods (L2P, DualPrompt, and S-Prompts), we achieve significant improvement on \textit{Transfer} metric, which shows that our DIKI mechanism can effectively inject new information to the frozen backbone without interfering with pre-trained knowledge. 

We also conduct experiments on the 16-shot MTIL-FS benchmark. Abbreviated results are shown in \cref{tab:fewshot} and the full table can be found in the supplementary materials. Since we only update a small amount of parameters, we gain more improvement over ZSCL compared to full parameter training. In addition, with minimal noise introduced, our fully residual IKI design demonstrates enhanced competitiveness when training data is deficient, compared to other interruptive prompt tuning methods.

\begin{figure}[t]
\begin{minipage}[!t]{0.49\textwidth}
    \centering
    \includegraphics[width=\linewidth]{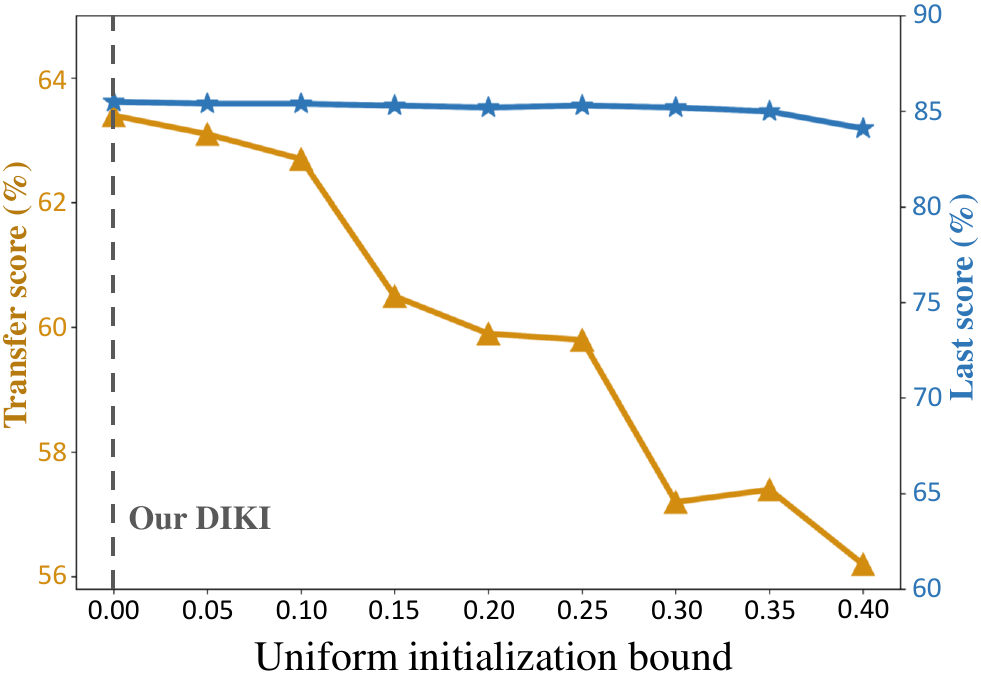} 
    \caption{Transfer and Last scores (\%) with different uniform initialization bounds for residual attention parameters on MTIL benchmark. A larger initialization value will not affect the final accuracy (Last score), but could have a severe adverse impact on the model's zero-shot ability, due to the random noise introduced into the pre-trained model.}
    \label{fig:diff_init_ablation}
\end{minipage}%
\hspace{0.02\textwidth}
\begin{minipage}[!t]{0.49\textwidth}
    \centering
    \includegraphics[width=\linewidth]{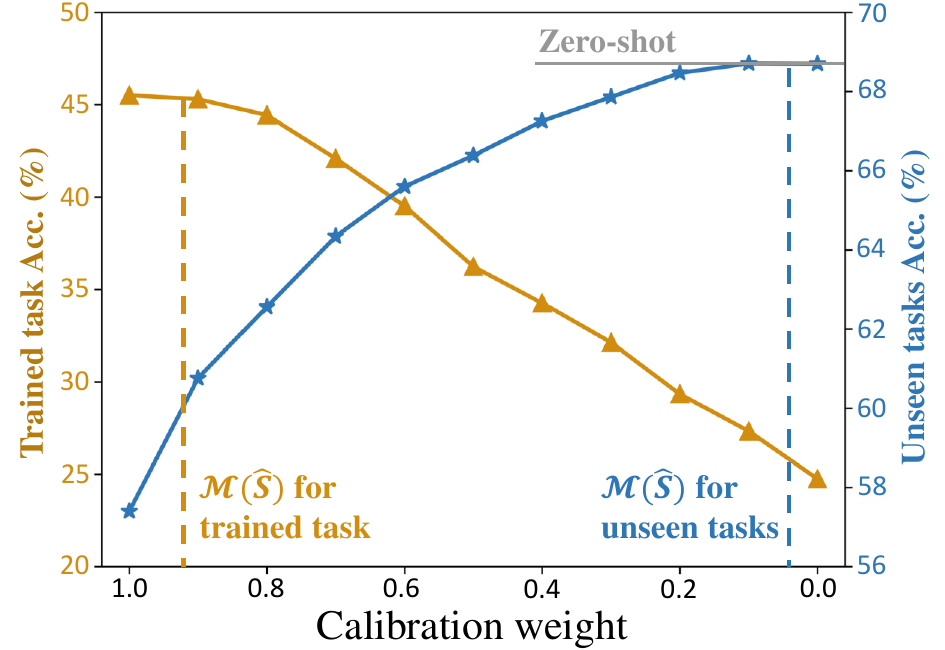} 
    \caption{Demonstration of the effect of our distribution-aware integration calibration. We evaluate the model, which is only trained on the first task of MTIL, on the trained task and unseen tasks, with \textbf{manually assigned} calibration weights. Fixed larger weights maintain high accuracy on trained task while lose zero-shot ability, and vice versa. Our DIKI tailors weight for different samples during inference time.}
    \label{fig:attention_weight_ablation}
\end{minipage}
\end{figure}

\subsection{Analysis}
\label{sec:analysis}

\textbf{Ablation study.}
We ablate our proposed modules of DIKI on MTIL benchmark in \cref{tab:main_ablation}. Firstly we consider \textit{Transfer} score (i.e. zero-shot ability): from the first two rows, it can be seen that the zero-initialization mechanism is ineffective to prompt tuning methods, because they can still disturb the pre-trained knowledge by softmax function inside the attention calculation. However, with our residual attention design, the effect of zero-initialization is activated. They can work together to avoid introducing irrelevant information to the frozen backbone. Thanks to the fully residual property, distribution-aware calibration can be exploited to further boost performance by identifying unseen distributions.

Considering the \textit{Last} metric, our interference-free mechanism stores more task knowledge because of its clear information injection process, thus enhancing the last state accuracy. However since our zero-initialization and distribution-aware calibration are designed to improve the retention of pre-trained knowledge, the addition of them does not result in an increase on \textit{Last} score.

\noindent\textbf{Effect of zero-initialized residual attention.}
To demonstrate the effect of our zero-initialization paradigm, we conduct experiments for different distributed initialization strategies on the MTIL benchmark, as shown in \cref{fig:diff_init_ablation}. Following previous common practice \cite{zhou2022learning,zhou2022conditional}, we choose uniform distribution with different bounds to initialize our trainable $K_r$ and $V_r$ in \cref{eq:res_attn}. Results show that with different initialization values, the model can achieve constant final performance after being trained on all tasks (\textit{Last} score keeps invariant). However, as the initialization bound increases, model's zero-shot ability degenerates due to the noise introduced by random initialization (\textit{Transfer} score is decreasing).

\noindent\textbf{Effect of distribution aware calibration.}
To demonstrate our calibration technique, we conduct experiments with manually set calibration weights. Specifically, we train the model exclusively on the first task of MTIL (Aircraft \cite{maji2013fine} dataset) and test it on all tasks, including trained and unseen datasets. Here we replace $\mathcal{M}(\hat{S})$ in \cref{eq:final_output} with fixed values, as shown in \cref{fig:attention_weight_ablation}. When the weight is set to 1.0, which means full use of newly learned knowledge, the trained task accuracy is maximized while the vital zero-shot ability is interfered with. Conversely, as weight decreases, the zero-shot capability returns, while trained task accuracy decreases due to the reduced incorporation of new knowledge.

Our distribution-aware attention calibration tailors appropriate weights for different inference samples by the distribution modeling, allocating higher/lower weights to samples from learned/unseen domains. It alleviates the need to select a ``balance point'' which compromises overall performance.


\begin{wraptable}[25]{r}{0.5\columnwidth}
    \centering
    \setlength{\belowcaptionskip}{1mm}
    \vspace{-11mm}
    \caption{Results of CIL task on the 10-split CIFAR-100 dataset. We replace the prepending mechanism used in previous prompt-based CIL methods with our IKI strategy. Its residual property facilitates knowledge acquisition and reduces noise, enhancing existing works plug-and-play.}
    \resizebox{0.5\columnwidth}{!}{
    \fontsize{8pt}{10pt}\selectfont
    \begin{tabular}{y{70}|x{60}x{60}}
        \toprule
        Method & Avg. Acc ($\uparrow$) & Forgetting ($\downarrow$) \\
        \midrule
        L2P~\cite{wang2022learning} & 83.86$\pm$0.28 & 7.35±0.38 \\
        \rowcolor{mygray} \quad + IKI & 84.61$\pm$0.20 & 7.28±0.31 \\
        \midrule
        DualPrompt~\cite{wang2022dualprompt} & 86.51$\pm$0.33 & 5.16$\pm$0.09 \\
        \rowcolor{mygray} \quad + IKI & \textbf{88.77$\pm$0.25} & 4.38$\pm$0.13 \\
        \midrule
        CODA-P~\cite{smith2023coda}  & 86.25$\pm$0.74 & 5.02$\pm$0.41 \\
        \rowcolor{mygray} \quad + IKI & 87.17$\pm$0.35 & \textbf{3.95$\pm$0.11} \\
        \bottomrule
    \end{tabular}
    }
    \label{tab:CIL}

    \vspace{1mm}
    
    \centering
    \caption{Training costs comparisons. ``GPU Mem.'' denotes the training requirement, and ``\# Ref img'' is the number of extra images used in the training stage except the continual training set. \textbf{-} means no extra data needed. We achieve higher performance with lower training costs.}
    \resizebox{0.5\columnwidth}{!}{
    \fontsize{8pt}{10pt}\selectfont
        \begin{tabular}{x{40}|x{38}x{26}x{48}x{42}}
            \toprule
             Method &  \# Param. & Time & GPU Mem. & \# Ref img \\
             \midrule
             ZSCL~\cite{wang2022learning} & 211 M & 11.3 h & 96 GB & 100k \\
             \rowcolor{mygray} DIKI & 1.8 M & 2.3 h & 24 GB & \textbf{-} \\
            \bottomrule
        \end{tabular}
    }
    \label{tab:computation_comp}
\end{wraptable}

\noindent\textbf{Effect of IKI on CIL.}
To validate the universality of the proposed IKI, we evaluate it on the conventional Class Incremental Learning (CIL) task. Specifically, IKI is integrated into existing prompt-based CIL methods, serving as a replacement for their original prepending mechanisms. Experiments are conducted on the 10-split CIFAR-100 dataset following the common protocol \cite{wang2022dualprompt,wang2022learning}, as shown in \cref{tab:CIL}. IKI explicitly formulates a knowledge injection process, thus boosting the average accuracy by achieving superior performance on each task. For the forgetting metric, result of L2P \cite{wang2022dualprompt} remains comparable due to the absence of shared information across tasks. Conversely, for methods with shared prompts (DualPrompt \cite{wang2022learning} and CODA-P \cite{smith2023coda}), our non-interference attention mechanism facilitates the knowledge shareability and alleviates the forgetting problem.

\noindent\textbf{Training cost analysis.}
We compare the computational requirement of our DIKI and previous state-of-the-art method ZSCL \cite{wang2022learning} in \cref{tab:computation_comp}. Benefiting from our parameter efficient framework, the training process of DIKI only lasts 2.3 hours on a single GPU, while ZSCL requires 4 GPUs, nearly half a day for training, and extra 100k images to perform distillation. With a much faster model adaptation speed, our method can be more effective and adoptable in tackling real-world continual learning problems.

\begin{figure}[t]
    \centering
     \setlength{\abovecaptionskip}{2mm}
    \includegraphics[width=\linewidth]{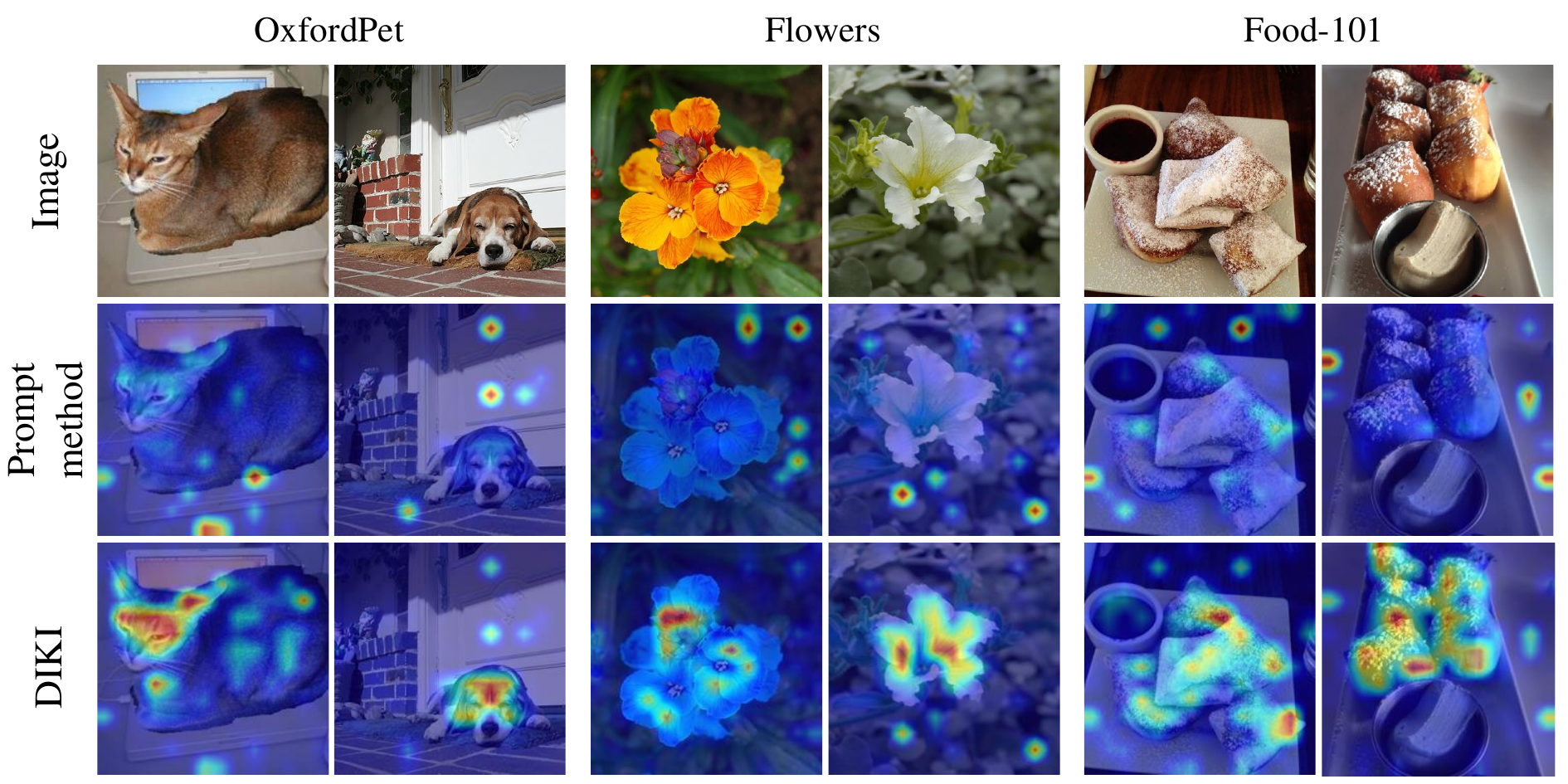} 
    \caption{Heatmap visualization comparisons. We employ Grad-CAM \cite{selvaraju2017grad} to evaluate the model, which only has been trained on Aircraft \cite{maji2013fine}, across unseen datasets OxfordPet \cite{parkhi2012cats}, Flowers \cite{nilsback2008automated} and Food-101 \cite{bossard2014food}. It demonstrates that the commonly used prompt-based methods introduce noise into the model, thus resulting in forward forgetting issue and model degradation. Our DIKI implants new knowledge in a fully residual manner, optimizing the retention of pre-trained knowledge.}
    \label{fig:grad_cam}
    \vspace{-2mm}
\end{figure}

\noindent\textbf{Qualitative visualization results.}
We implement Grad-CAM \cite{selvaraju2017grad} on the attention maps of the CLIP visual encoder, following the practice used in \cite{shen2021much}, as depicted in \cref{fig:grad_cam}. Specifically, we load the model which is only trained on the first dataset Aircraft \cite{maji2013fine} of MTIL benchmark, and test it on several subsequent unseen datasets. We observe that the vanilla prompting way (employed by L2P, Dualprompt, and S-Prompts) interferes with pre-trained knowledge and undermines the zero-shot ability. However, with utilizing our DIKI, the generalization ability acquired during pre-training is preserved.


\section{Conclusions}


This study introduced Distribution-aware Interference-free Knowledge Integration (DIKI) mechanism for domain-class incremental learning. DIKI preserves the pre-trained knowledge of VLMs while effectively implanting new task information, without heavy computation and external data.
DIKI infuses new knowledge into a frozen backbone in a fully residual manner, effectively mitigating the forward forgetting issue. A distribution-aware integration calibration technique is also integrated, which controls the information injection for data from unseen distributions.
Experiments show that DIKI surpasses the previous SOTA method with only 0.86\% trainable parameters.

\noindent\textbf{Limitations and future work.}
Our DIKI follows a task-specific tuning paradigm, where the training on different tasks is independent. Although some recent CIL research works have verified the effect of sharing knowledge across tasks \cite{wang2022dualprompt,smith2023coda}, we find these solutions are impractical within the DCIL context. Experiments are conducted in the supplementary materials. We attribute this to the significant domain gap among DCIL datasets, which hinders the shareability of knowledge from different tasks. Future works could explore suitable knowledge-sharing strategies tailored to the DCIL problems.

%
%
\bibliographystyle{splncs04}
\bibliography{main}

\begin{thebibliography}{10}
\providecommand{\url}[1]{\texttt{#1}}
\providecommand{\urlprefix}{URL }
\providecommand{\doi}[1]{https://doi.org/#1}

\bibitem{ahn2019uncertainty}
Ahn, H., Cha, S., Lee, D., Moon, T.: Uncertainty-based continual learning with
  adaptive regularization. Advances in neural information processing systems
  \textbf{32} (2019)

\bibitem{aljundi2018memory}
Aljundi, R., Babiloni, F., Elhoseiny, M., Rohrbach, M., Tuytelaars, T.: Memory
  aware synapses: Learning what (not) to forget. In: Proceedings of the
  European conference on computer vision (ECCV). pp. 139--154 (2018)

\bibitem{bossard2014food}
Bossard, L., Guillaumin, M., Van~Gool, L.: Food-101--mining discriminative
  components with random forests. In: Computer Vision--ECCV 2014: 13th European
  Conference, Zurich, Switzerland, September 6-12, 2014, Proceedings, Part VI
  13. pp. 446--461. Springer (2014)

\bibitem{bowman2023carte}
Bowman, B., Achille, A., Zancato, L., Trager, M., Perera, P., Paolini, G.,
  Soatto, S.: a-la-carte prompt tuning (apt): Combining distinct data via
  composable prompting. In: Proceedings of the IEEE/CVF Conference on Computer
  Vision and Pattern Recognition. pp. 14984--14993 (2023)

\bibitem{chen2022adaptformer}
Chen, S., Ge, C., Tong, Z., Wang, J., Song, Y., Wang, J., Luo, P.: Adaptformer:
  Adapting vision transformers for scalable visual recognition. Advances in
  Neural Information Processing Systems  \textbf{35},  16664--16678 (2022)

\bibitem{cimpoi2014describing}
Cimpoi, M., Maji, S., Kokkinos, I., Mohamed, S., Vedaldi, A.: Describing
  textures in the wild. In: Proceedings of the IEEE conference on computer
  vision and pattern recognition. pp. 3606--3613 (2014)

\bibitem{de2021continual}
De~Lange, M., Aljundi, R., Masana, M., Parisot, S., Jia, X., Leonardis, A.,
  Slabaugh, G., Tuytelaars, T.: A continual learning survey: Defying forgetting
  in classification tasks. IEEE transactions on pattern analysis and machine
  intelligence  \textbf{44}(7),  3366--3385 (2021)

\bibitem{deng2009imagenet}
Deng, J., Dong, W., Socher, R., Li, L.J., Li, K., Fei-Fei, L.: Imagenet: A
  large-scale hierarchical image database. In: 2009 IEEE conference on computer
  vision and pattern recognition. pp. 248--255. Ieee (2009)

\bibitem{deng2012mnist}
Deng, L.: The mnist database of handwritten digit images for machine learning
  research [best of the web]. IEEE signal processing magazine  \textbf{29}(6),
  141--142 (2012)

\bibitem{dhariwal2021diffusion}
Dhariwal, P., Nichol, A.: Diffusion models beat gans on image synthesis.
  Advances in neural information processing systems  \textbf{34},  8780--8794
  (2021)

\bibitem{ding2022don}
Ding, Y., Liu, L., Tian, C., Yang, J., Ding, H.: Don't stop learning: Towards
  continual learning for the clip model. arXiv preprint arXiv:2207.09248
  (2022)

\bibitem{dosovitskiy2020image}
Dosovitskiy, A., Beyer, L., Kolesnikov, A., Weissenborn, D., Zhai, X.,
  Unterthiner, T., Dehghani, M., Minderer, M., Heigold, G., Gelly, S., et~al.:
  An image is worth 16x16 words: Transformers for image recognition at scale.
  arXiv preprint arXiv:2010.11929  (2020)

\bibitem{douillard2022dytox}
Douillard, A., Ram{\'e}, A., Couairon, G., Cord, M.: Dytox: Transformers for
  continual learning with dynamic token expansion. In: Proceedings of the
  IEEE/CVF Conference on Computer Vision and Pattern Recognition. pp.
  9285--9295 (2022)

\bibitem{fang2024real}
Fang, C., He, C., Xiao, F., Zhang, Y., Tang, L., Zhang, Y., Li, K., Li, X.:
  Real-world image dehazing with coherence-based label generator and
  cooperative unfolding network. arXiv preprint arXiv:2406.07966  (2024)

\bibitem{fei2004learning}
Fei-Fei, L., Fergus, R., Perona, P.: Learning generative visual models from few
  training examples: An incremental bayesian approach tested on 101 object
  categories. In: 2004 conference on computer vision and pattern recognition
  workshop. pp. 178--178. IEEE (2004)

\bibitem{gao2023clip}
Gao, P., Geng, S., Zhang, R., Ma, T., Fang, R., Zhang, Y., Li, H., Qiao, Y.:
  Clip-adapter: Better vision-language models with feature adapters.
  International Journal of Computer Vision pp. 1--15 (2023)

\bibitem{he2023reti}
He, C., Fang, C., Zhang, Y., Li, K., Tang, L., You, C., Xiao, F., Guo, Z., Li,
  X.: Reti-diff: Illumination degradation image restoration with retinex-based
  latent diffusion model. arXiv preprint arXiv:2311.11638  (2023)

\bibitem{he2023camouflaged}
He, C., Li, K., Zhang, Y., Tang, L., Zhang, Y., Guo, Z., Li, X.: Camouflaged
  object detection with feature decomposition and edge reconstruction. In:
  CVPR. pp. 22046--22055 (2023)

\bibitem{he2023weaklysupervised}
He, C., Li, K., Zhang, Y., Xu, G., Tang, L.: Weakly-supervised concealed object
  segmentation with sam-based pseudo labeling and multi-scale feature grouping.
  NeurIPS  (2024)

\bibitem{he2023strategic}
He, C., Li, K., Zhang, Y., Zhang, Y., Guo, Z., Li, X.: Strategic preys make
  acute predators: Enhancing camouflaged object detectors by generating
  camouflaged objects. In: ICLR (2024)

\bibitem{he2024diffusion}
He, C., Shen, Y., Fang, C., Xiao, F., Tang, L., Zhang, Y., Zuo, W., Guo, Z.,
  Li, X.: Diffusion models in low-level vision: A survey. arXiv preprint
  arXiv:2406.11138  (2024)

\bibitem{he2016deep}
He, K., Zhang, X., Ren, S., Sun, J.: Deep residual learning for image
  recognition. In: Proceedings of the IEEE conference on computer vision and
  pattern recognition. pp. 770--778 (2016)

\bibitem{hegde2023clip}
Hegde, D., Valanarasu, J.M.J., Patel, V.: Clip goes 3d: Leveraging prompt
  tuning for language grounded 3d recognition. In: Proceedings of the IEEE/CVF
  International Conference on Computer Vision. pp. 2028--2038 (2023)

\bibitem{helber2019eurosat}
Helber, P., Bischke, B., Dengel, A., Borth, D.: Eurosat: A novel dataset and
  deep learning benchmark for land use and land cover classification. IEEE
  Journal of Selected Topics in Applied Earth Observations and Remote Sensing
  \textbf{12}(7),  2217--2226 (2019)

\bibitem{houlsby2019parameter}
Houlsby, N., Giurgiu, A., Jastrzebski, S., Morrone, B., De~Laroussilhe, Q.,
  Gesmundo, A., Attariyan, M., Gelly, S.: Parameter-efficient transfer learning
  for nlp. In: International Conference on Machine Learning. pp. 2790--2799.
  PMLR (2019)

\bibitem{hu2021lora}
Hu, E.J., Shen, Y., Wallis, P., Allen-Zhu, Z., Li, Y., Wang, S., Wang, L.,
  Chen, W.: Lora: Low-rank adaptation of large language models. arXiv preprint
  arXiv:2106.09685  (2021)

\bibitem{hu2023pop}
Hu, Z., Lyu, J., Gao, D., Vasconcelos, N.: Pop: Prompt of prompts for continual
  learning. arXiv preprint arXiv:2306.08200  (2023)

\bibitem{isele2018selective}
Isele, D., Cosgun, A.: Selective experience replay for lifelong learning. In:
  Proceedings of the AAAI Conference on Artificial Intelligence. vol.~32 (2018)

\bibitem{jia2021scaling}
Jia, C., Yang, Y., Xia, Y., Chen, Y.T., Parekh, Z., Pham, H., Le, Q., Sung,
  Y.H., Li, Z., Duerig, T.: Scaling up visual and vision-language
  representation learning with noisy text supervision. In: International
  conference on machine learning. pp. 4904--4916. PMLR (2021)

\bibitem{jia2022visual}
Jia, M., Tang, L., Chen, B.C., Cardie, C., Belongie, S., Hariharan, B., Lim,
  S.N.: Visual prompt tuning. In: European Conference on Computer Vision. pp.
  709--727. Springer (2022)

\bibitem{jie2023fact}
Jie, S., Deng, Z.H.: Fact: Factor-tuning for lightweight adaptation on vision
  transformer. In: Proceedings of the AAAI Conference on Artificial
  Intelligence. vol.~37, pp. 1060--1068 (2023)

\bibitem{ju2022prompting}
Ju, C., Han, T., Zheng, K., Zhang, Y., Xie, W.: Prompting visual-language
  models for efficient video understanding. In: European Conference on Computer
  Vision. pp. 105--124. Springer (2022)

\bibitem{khan2023introducing}
Khan, M.G.Z.A., Naeem, M.F., Van~Gool, L., Stricker, D., Tombari, F., Afzal,
  M.Z.: Introducing language guidance in prompt-based continual learning. In:
  Proceedings of the IEEE/CVF International Conference on Computer Vision. pp.
  11463--11473 (2023)

\bibitem{khattak2023maple}
Khattak, M.U., Rasheed, H., Maaz, M., Khan, S., Khan, F.S.: Maple: Multi-modal
  prompt learning. In: Proceedings of the IEEE/CVF Conference on Computer
  Vision and Pattern Recognition. pp. 19113--19122 (2023)

\bibitem{kirkpatrick2017overcoming}
Kirkpatrick, J., Pascanu, R., Rabinowitz, N., Veness, J., Desjardins, G., Rusu,
  A.A., Milan, K., Quan, J., Ramalho, T., Grabska-Barwinska, A., et~al.:
  Overcoming catastrophic forgetting in neural networks. Proceedings of the
  national academy of sciences  \textbf{114}(13),  3521--3526 (2017)

\bibitem{krause20133d}
Krause, J., Stark, M., Deng, J., Fei-Fei, L.: 3d object representations for
  fine-grained categorization. In: Proceedings of the IEEE international
  conference on computer vision workshops. pp. 554--561 (2013)

\bibitem{krizhevsky2009learning}
Krizhevsky, A., Hinton, G., et~al.: Learning multiple layers of features from
  tiny images  (2009)

\bibitem{lai2024lisa}
Lai, X., Tian, Z., Chen, Y., Li, Y., Yuan, Y., Liu, S., Jia, J.: Lisa:
  Reasoning segmentation via large language model. In: Proceedings of the
  IEEE/CVF Conference on Computer Vision and Pattern Recognition. pp.
  9579--9589 (2024)

\bibitem{lai2021semi}
Lai, X., Tian, Z., Jiang, L., Liu, S., Zhao, H., Wang, L., Jia, J.:
  Semi-supervised semantic segmentation with directional context-aware
  consistency. In: Proceedings of the IEEE/CVF Conference on Computer Vision
  and Pattern Recognition. pp. 1205--1214 (2021)

\bibitem{li2022blip}
Li, J., Li, D., Xiong, C., Hoi, S.: Blip: Bootstrapping language-image
  pre-training for unified vision-language understanding and generation. In:
  International Conference on Machine Learning. pp. 12888--12900. PMLR (2022)

\bibitem{li2021prefix}
Li, X.L., Liang, P.: Prefix-tuning: Optimizing continuous prompts for
  generation. arXiv preprint arXiv:2101.00190  (2021)

\bibitem{li2019learn}
Li, X., Zhou, Y., Wu, T., Socher, R., Xiong, C.: Learn to grow: A continual
  structure learning framework for overcoming catastrophic forgetting. In:
  International Conference on Machine Learning. pp. 3925--3934. PMLR (2019)

\bibitem{li2017learning}
Li, Z., Hoiem, D.: Learning without forgetting. IEEE transactions on pattern
  analysis and machine intelligence  \textbf{40}(12),  2935--2947 (2017)

\bibitem{liu2023pre}
Liu, P., Yuan, W., Fu, J., Jiang, Z., Hayashi, H., Neubig, G.: Pre-train,
  prompt, and predict: A systematic survey of prompting methods in natural
  language processing. ACM Computing Surveys  \textbf{55}(9),  1--35 (2023)

\bibitem{lopez2017gradient}
Lopez-Paz, D., Ranzato, M.: Gradient episodic memory for continual learning.
  Advances in neural information processing systems  \textbf{30} (2017)

\bibitem{maji2013fine}
Maji, S., Rahtu, E., Kannala, J., Blaschko, M., Vedaldi, A.: Fine-grained
  visual classification of aircraft. arXiv preprint arXiv:1306.5151  (2013)

\bibitem{mallya2018packnet}
Mallya, A., Lazebnik, S.: Packnet: Adding multiple tasks to a single network by
  iterative pruning. In: Proceedings of the IEEE conference on Computer Vision
  and Pattern Recognition. pp. 7765--7773 (2018)

\bibitem{nilsback2008automated}
Nilsback, M.E., Zisserman, A.: Automated flower classification over a large
  number of classes. In: 2008 Sixth Indian conference on computer vision,
  graphics \& image processing. pp. 722--729. IEEE (2008)

\bibitem{parkhi2012cats}
Parkhi, O.M., Vedaldi, A., Zisserman, A., Jawahar, C.: Cats and dogs. In: 2012
  IEEE conference on computer vision and pattern recognition. pp. 3498--3505.
  IEEE (2012)

\bibitem{peng2023hierarchical}
Peng, B., Tian, Z., Wu, X., Wang, C., Liu, S., Su, J., Jia, J.: Hierarchical
  dense correlation distillation for few-shot segmentation. In: Proceedings of
  the IEEE/CVF Conference on Computer Vision and Pattern Recognition. pp.
  23641--23651 (2023)

\bibitem{prabhu2020gdumb}
Prabhu, A., Torr, P.H., Dokania, P.K.: Gdumb: A simple approach that questions
  our progress in continual learning. In: Computer Vision--ECCV 2020: 16th
  European Conference, Glasgow, UK, August 23--28, 2020, Proceedings, Part II
  16. pp. 524--540. Springer (2020)

\bibitem{qian2023decouple}
Qian, Z., Wang, X., Duan, X., Qin, P., Li, Y., Zhu, W.: Decouple before
  interact: Multi-modal prompt learning for continual visual question
  answering. In: Proceedings of the IEEE/CVF International Conference on
  Computer Vision. pp. 2953--2962 (2023)

\bibitem{radford2021learning}
Radford, A., Kim, J.W., Hallacy, C., Ramesh, A., Goh, G., Agarwal, S., Sastry,
  G., Askell, A., Mishkin, P., Clark, J., et~al.: Learning transferable visual
  models from natural language supervision. In: International conference on
  machine learning. pp. 8748--8763. PMLR (2021)

\bibitem{rao2019continual}
Rao, D., Visin, F., Rusu, A., Pascanu, R., Teh, Y.W., Hadsell, R.: Continual
  unsupervised representation learning. Advances in neural information
  processing systems  \textbf{32} (2019)

\bibitem{rebuffi2017icarl}
Rebuffi, S.A., Kolesnikov, A., Sperl, G., Lampert, C.H.: icarl: Incremental
  classifier and representation learning. In: Proceedings of the IEEE
  conference on Computer Vision and Pattern Recognition. pp. 2001--2010 (2017)

\bibitem{rolnick2019experience}
Rolnick, D., Ahuja, A., Schwarz, J., Lillicrap, T., Wayne, G.: Experience
  replay for continual learning. Advances in Neural Information Processing
  Systems  \textbf{32} (2019)

\bibitem{saharia2022photorealistic}
Saharia, C., Chan, W., Saxena, S., Li, L., Whang, J., Denton, E.L.,
  Ghasemipour, K., Gontijo~Lopes, R., Karagol~Ayan, B., Salimans, T., et~al.:
  Photorealistic text-to-image diffusion models with deep language
  understanding. Advances in Neural Information Processing Systems
  \textbf{35},  36479--36494 (2022)

\bibitem{selvaraju2017grad}
Selvaraju, R.R., Cogswell, M., Das, A., Vedantam, R., Parikh, D., Batra, D.:
  Grad-cam: Visual explanations from deep networks via gradient-based
  localization. In: Proceedings of the IEEE international conference on
  computer vision. pp. 618--626 (2017)

\bibitem{shen2021much}
Shen, S., Li, L.H., Tan, H., Bansal, M., Rohrbach, A., Chang, K.W., Yao, Z.,
  Keutzer, K.: How much can clip benefit vision-and-language tasks? arXiv
  preprint arXiv:2107.06383  (2021)

\bibitem{shin2017continual}
Shin, H., Lee, J.K., Kim, J., Kim, J.: Continual learning with deep generative
  replay. Advances in neural information processing systems  \textbf{30} (2017)

\bibitem{smith2023construct}
Smith, J.S., Cascante-Bonilla, P., Arbelle, A., Kim, D., Panda, R., Cox, D.,
  Yang, D., Kira, Z., Feris, R., Karlinsky, L.: Construct-vl: Data-free
  continual structured vl concepts learning. In: Proceedings of the IEEE/CVF
  Conference on Computer Vision and Pattern Recognition. pp. 14994--15004
  (2023)

\bibitem{smith2023coda}
Smith, J.S., Karlinsky, L., Gutta, V., Cascante-Bonilla, P., Kim, D., Arbelle,
  A., Panda, R., Feris, R., Kira, Z.: Coda-prompt: Continual decomposed
  attention-based prompting for rehearsal-free continual learning. In:
  Proceedings of the IEEE/CVF Conference on Computer Vision and Pattern
  Recognition. pp. 11909--11919 (2023)

\bibitem{sohn2023visual}
Sohn, K., Chang, H., Lezama, J., Polania, L., Zhang, H., Hao, Y., Essa, I.,
  Jiang, L.: Visual prompt tuning for generative transfer learning. In:
  Proceedings of the IEEE/CVF Conference on Computer Vision and Pattern
  Recognition. pp. 19840--19851 (2023)

\bibitem{tang2023consistency}
Tang, L., Li, K., He, C., Zhang, Y., Li, X.: Consistency regularization for
  generalizable source-free domain adaptation. In: Proceedings of the IEEE/CVF
  International Conference on Computer Vision. pp. 4323--4333 (2023)

\bibitem{tang2023source}
Tang, L., Li, K., He, C., Zhang, Y., Li, X.: Source-free domain adaptive fundus
  image segmentation with class-balanced mean teacher. In: International
  Conference on Medical Image Computing and Computer-Assisted Intervention. pp.
  684--694. Springer (2023)

\bibitem{tian2022generalized}
Tian, Z., Lai, X., Jiang, L., Liu, S., Shu, M., Zhao, H., Jia, J.: Generalized
  few-shot semantic segmentation. In: Proceedings of the IEEE/CVF Conference on
  Computer Vision and Pattern Recognition. pp. 11563--11572 (2022)

\bibitem{tian2019learning}
Tian, Z., Shu, M., Lyu, P., Li, R., Zhou, C., Shen, X., Jia, J.: Learning
  shape-aware embedding for scene text detection. In: Proceedings of the
  IEEE/CVF conference on computer vision and pattern recognition. pp.
  4234--4243 (2019)

\bibitem{tian2020prior}
Tian, Z., Zhao, H., Shu, M., Yang, Z., Li, R., Jia, J.: Prior guided feature
  enrichment network for few-shot segmentation. IEEE transactions on pattern
  analysis and machine intelligence  \textbf{44}(2),  1050--1065 (2020)

\bibitem{vaswani2017attention}
Vaswani, A., Shazeer, N., Parmar, N., Uszkoreit, J., Jones, L., Gomez, A.N.,
  Kaiser, {\L}., Polosukhin, I.: Attention is all you need. Advances in neural
  information processing systems  \textbf{30} (2017)

\bibitem{van2019three}
Van~de Ven, G.M., Tolias, A.S.: Three scenarios for continual learning. arXiv
  preprint arXiv:1904.07734  (2019)

\bibitem{wang2020kadapter}
Wang, R., Tang, D., Duan, N., Wei, Z., Huang, X., ji, J., Cao, G., Jiang, D.,
  Zhou, M.: K-adapter: Infusing knowledge into pre-trained models with adapters
  (2020)

\bibitem{wang2022s}
Wang, Y., Huang, Z., Hong, X.: S-prompts learning with pre-trained
  transformers: An occam’s razor for domain incremental learning. Advances in
  Neural Information Processing Systems  \textbf{35},  5682--5695 (2022)

\bibitem{wang2022dualprompt}
Wang, Z., Zhang, Z., Ebrahimi, S., Sun, R., Zhang, H., Lee, C.Y., Ren, X., Su,
  G., Perot, V., Dy, J., et~al.: Dualprompt: Complementary prompting for
  rehearsal-free continual learning. In: European Conference on Computer
  Vision. pp. 631--648. Springer (2022)

\bibitem{wang2022learning}
Wang, Z., Zhang, Z., Lee, C.Y., Zhang, H., Sun, R., Ren, X., Su, G., Perot, V.,
  Dy, J., Pfister, T.: Learning to prompt for continual learning. In:
  Proceedings of the IEEE/CVF Conference on Computer Vision and Pattern
  Recognition. pp. 139--149 (2022)

\bibitem{wortsman2022robust}
Wortsman, M., Ilharco, G., Kim, J.W., Li, M., Kornblith, S., Roelofs, R.,
  Lopes, R.G., Hajishirzi, H., Farhadi, A., Namkoong, H., et~al.: Robust
  fine-tuning of zero-shot models. In: Proceedings of the IEEE/CVF Conference
  on Computer Vision and Pattern Recognition. pp. 7959--7971 (2022)

\bibitem{xiao2010sun}
Xiao, J., Hays, J., Ehinger, K.A., Oliva, A., Torralba, A.: Sun database:
  Large-scale scene recognition from abbey to zoo. In: 2010 IEEE computer
  society conference on computer vision and pattern recognition. pp.
  3485--3492. IEEE (2010)

\bibitem{yang2024v}
Yang, J., Ding, R., Brown, E., Qi, X., Xie, S.: V-irl: Grounding virtual
  intelligence in real life. arXiv preprint arXiv:2402.03310  (2024)

\bibitem{yang2024unified}
Yang, S., Tian, Z., Jiang, L., Jia, J.: Unified language-driven zero-shot
  domain adaptation. In: Proceedings of the IEEE/CVF Conference on Computer
  Vision and Pattern Recognition. pp. 23407--23415 (2024)

\bibitem{yang2024exploring}
Yang, S., Wu, J., Liu, J., Li, X., Zhang, Q., Pan, M., Gan, Y., Chen, Z.,
  Zhang, S.: Exploring sparse visual prompt for domain adaptive dense
  prediction. In: Proceedings of the AAAI Conference on Artificial
  Intelligence. vol.~38, pp. 16334--16342 (2024)

\bibitem{yao2021filip}
Yao, L., Huang, R., Hou, L., Lu, G., Niu, M., Xu, H., Liang, X., Li, Z., Jiang,
  X., Xu, C.: Filip: Fine-grained interactive language-image pre-training.
  arXiv preprint arXiv:2111.07783  (2021)

\bibitem{yoon2017lifelong}
Yoon, J., Yang, E., Lee, J., Hwang, S.J.: Lifelong learning with dynamically
  expandable networks. arXiv preprint arXiv:1708.01547  (2017)

\bibitem{zhai2022lit}
Zhai, X., Wang, X., Mustafa, B., Steiner, A., Keysers, D., Kolesnikov, A.,
  Beyer, L.: Lit: Zero-shot transfer with locked-image text tuning. In:
  Proceedings of the IEEE/CVF Conference on Computer Vision and Pattern
  Recognition. pp. 18123--18133 (2022)

\bibitem{zhang2020class}
Zhang, J., Zhang, J., Ghosh, S., Li, D., Tasci, S., Heck, L., Zhang, H., Kuo,
  C.C.J.: Class-incremental learning via deep model consolidation. In:
  Proceedings of the IEEE/CVF Winter Conference on Applications of Computer
  Vision. pp. 1131--1140 (2020)

\bibitem{zhang2023llama}
Zhang, R., Han, J., Zhou, A., Hu, X., Yan, S., Lu, P., Li, H., Gao, P., Qiao,
  Y.: Llama-adapter: Efficient fine-tuning of language models with zero-init
  attention. arXiv preprint arXiv:2303.16199  (2023)

\bibitem{zheng2023preventing}
Zheng, Z., Ma, M., Wang, K., Qin, Z., Yue, X., You, Y.: Preventing zero-shot
  transfer degradation in continual learning of vision-language models. arXiv
  preprint arXiv:2303.06628  (2023)

\bibitem{zhou2023learning}
Zhou, D.W., Zhang, Y., Ning, J., Ye, H.J., Zhan, D.C., Liu, Z.: Learning
  without forgetting for vision-language models. arXiv preprint
  arXiv:2305.19270  (2023)

\bibitem{zhou2024unihead}
Zhou, H., Yang, R., Zhang, Y., Duan, H., Huang, Y., Hu, R., Li, X., Zheng, Y.:
  Unihead: unifying multi-perception for detection heads. IEEE Transactions on
  Neural Networks and Learning Systems  (2024)

\bibitem{zhou2022conditional}
Zhou, K., Yang, J., Loy, C.C., Liu, Z.: Conditional prompt learning for
  vision-language models. In: Proceedings of the IEEE/CVF Conference on
  Computer Vision and Pattern Recognition. pp. 16816--16825 (2022)

\bibitem{zhou2022learning}
Zhou, K., Yang, J., Loy, C.C., Liu, Z.: Learning to prompt for vision-language
  models. International Journal of Computer Vision  \textbf{130}(9),
  2337--2348 (2022)

\end{thebibliography}


\begin{thebibliography}{10}
\providecommand{\url}[1]{\texttt{#1}}
\providecommand{\urlprefix}{URL }
\providecommand{\doi}[1]{https://doi.org/#1}

\bibitem{ding2022don}
Ding, Y., Liu, L., Tian, C., Yang, J., Ding, H.: Don't stop learning: Towards
  continual learning for the clip model. arXiv preprint arXiv:2207.09248
  (2022)

\bibitem{li2017learning}
Li, Z., Hoiem, D.: Learning without forgetting. IEEE transactions on pattern
  analysis and machine intelligence  \textbf{40}(12),  2935--2947 (2017)

\bibitem{rebuffi2017icarl}
Rebuffi, S.A., Kolesnikov, A., Sperl, G., Lampert, C.H.: icarl: Incremental
  classifier and representation learning. In: Proceedings of the IEEE
  conference on Computer Vision and Pattern Recognition. pp. 2001--2010 (2017)

\bibitem{smith2023coda}
Smith, J.S., Karlinsky, L., Gutta, V., Cascante-Bonilla, P., Kim, D., Arbelle,
  A., Panda, R., Feris, R., Kira, Z.: Coda-prompt: Continual decomposed
  attention-based prompting for rehearsal-free continual learning. In:
  Proceedings of the IEEE/CVF Conference on Computer Vision and Pattern
  Recognition. pp. 11909--11919 (2023)

\bibitem{wang2022s}
Wang, Y., Huang, Z., Hong, X.: S-prompts learning with pre-trained
  transformers: An occam’s razor for domain incremental learning. Advances in
  Neural Information Processing Systems  \textbf{35},  5682--5695 (2022)

\bibitem{wang2022dualprompt}
Wang, Z., Zhang, Z., Ebrahimi, S., Sun, R., Zhang, H., Lee, C.Y., Ren, X., Su,
  G., Perot, V., Dy, J., et~al.: Dualprompt: Complementary prompting for
  rehearsal-free continual learning. In: European Conference on Computer
  Vision. pp. 631--648. Springer (2022)

\bibitem{wang2022learning}
Wang, Z., Zhang, Z., Lee, C.Y., Zhang, H., Sun, R., Ren, X., Su, G., Perot, V.,
  Dy, J., Pfister, T.: Learning to prompt for continual learning. In:
  Proceedings of the IEEE/CVF Conference on Computer Vision and Pattern
  Recognition. pp. 139--149 (2022)

\bibitem{wortsman2022robust}
Wortsman, M., Ilharco, G., Kim, J.W., Li, M., Kornblith, S., Roelofs, R.,
  Lopes, R.G., Hajishirzi, H., Farhadi, A., Namkoong, H., et~al.: Robust
  fine-tuning of zero-shot models. In: Proceedings of the IEEE/CVF Conference
  on Computer Vision and Pattern Recognition. pp. 7959--7971 (2022)

\bibitem{zheng2023preventing}
Zheng, Z., Ma, M., Wang, K., Qin, Z., Yue, X., You, Y.: Preventing zero-shot
  transfer degradation in continual learning of vision-language models. arXiv
  preprint arXiv:2303.06628  (2023)

\bibitem{zhou2022learning}
Zhou, K., Yang, J., Loy, C.C., Liu, Z.: Learning to prompt for vision-language
  models. International Journal of Computer Vision  \textbf{130}(9),
  2337--2348 (2022)

\end{thebibliography}
\end{document}


\title{Supplementary Materials} 

\titlerunning{DIKI}

\author{
}

\authorrunning{L. Tang et al.}

\institute{
}

\maketitle

\section{Proof of the Initialization Strategy in Eq. (7)}

In Eq. (7), we stated that only values $V_r^{\text{init}}$ should be initialized to zero, while keys $K_r$ need to be random at the beginning. We argue that initializing both $K_r$ and $V_r$ to zero will result in $K_r$ remaining zero throughout the whole training process, and cause $V_r$ to degenerate into a matrix where all vectors are identical. Here, we provide a brief proof.

Recall the self-attention process for a single query vector $\bm{q}\in \mathbb{R}^{d}$, where $d$ is the model embedding dimension. Note that in this proof, subscripts $m$ and $i$ denote the corresponding vectors of a matrix, while $n$ and $j$ denote subscripts for the individual elements within a vector.

We first derive the attention vector $\bm{z}$ with
\begin{equation}
\label{eq:z_m}
    \bm{z}_m = \frac{\sum_{j=1}^d \bm{q}_j \bm{K}_{m,j}}{\sqrt{d}}, \quad \bm{z} \in \mathbb{R}^{l}
\end{equation}
where $\bm{K}\in \mathbb{R}^{l\times d}$ is the trainable key matrix, and the subscript represents taking the corresponding element. Then a softmax function will be applied to get normalized attention score $\bm{a}$:
\begin{equation}
\label{eq:a_m}
    \bm{a}_m = \frac{e^{\bm{z}_m}}{\sum_{j=1}^l e^{\bm{z}_j}}, \quad \bm{a} \in \mathbb{R}^{l}
\end{equation}

Finally, the layer output vector $\bm{o}$ of the input query $\bm{q}$ can be derived with
\begin{equation}
\label{eq:o_n}
    \bm{o}_n = \sum_{i=1}^l \bm{a}_i \bm{V}_{i,n}, \quad \bm{o} \in \mathbb{R}^{d}
\end{equation}
where $\bm{V}\in \mathbb{R}^{l\times d}$ is the trainable value matrix.

Now we prove our statement with these preliminaries.

\noindent \textbf{(1)} First we discuss the situation that both key $\bm{K}$ and value $\bm{V}$ matrices are initialized to zero. Here we omit the multi-layer design of transformers and focus on one single attention layer. Assume that we have ground truth for output vector $\bm{o}$, and then we can get the training loss $\mathcal{L}$. Then we can calculate the derivative of $\mathcal{L}$ with respect to $\bm{K}$ and $\bm{V}$.

(\textit{i}) The first training step.

Here we use the parenthesized superscript the denote the parameter after the corresponding training step. Before the first training step, we have
\begin{equation}
\label{eq:K_0_V_0}
    \bm{K}^{(0)} = \bm{V}^{(0)} = [\bm{0}]^{l\times d}
\end{equation}

For the derivative of $\mathcal{L}$ with respect to $\bm{K}_{m,n}^{(0)}$, we have
\begin{equation}
\label{eq:d_L_K}
\begin{split}
    \frac{\partial \mathcal{L}}{\partial \bm{K}_{m,n}^{(0)}} &= \sum_{j=1}^d \frac{\partial \mathcal{L}}{\partial \bm{o}_j^{(0)}}\frac{\partial \bm{o}_j^{(0)}}{\partial \bm{K}_{m,n}^{(0)}} \\
    &= \sum_{j=1}^d \frac{\partial \mathcal{L}}{\partial \bm{o}_j^{(0)}}\left( \sum_{i=1}^l \frac{\partial \bm{o}_j^{(0)}}{\partial \bm{a}_{i}^{(0)}} \frac{\partial \bm{a}_i^{(0)}}{\partial \bm{K}_{m,n}^{(0)}}\right) \\
    &= \sum_{j=1}^d \frac{\partial \mathcal{L}}{\partial \bm{o}_j^{(0)}}\left( \sum_{i=1}^l \frac{\partial \bm{o}_j^{(0)}}{\partial \bm{a}_{i}^{(0)}} \frac{\partial \bm{a}_i^{(0)}}{\partial \bm{z}_{m}^{(0)}} \frac{\partial \bm{z}_m^{(0)}}{\partial \bm{K}_{m,n}^{(0)}}\right) \\
\end{split}
\end{equation}

Based on \cref{eq:z_m,eq:o_n}, we know
\begin{equation}
\label{eq:d_o_a_z_k}
    \frac{\partial \bm{o}_j^{(0)}}{\partial \bm{a}_{i}^{(0)}} = \bm{V}_{i,j}^{(0)}, \quad \frac{\partial \bm{z}_m^{(0)}}{\partial \bm{K}_{m,n}^{(0)}} = \bm{q}_n
\end{equation}
and $\frac{\partial \mathcal{L}}{\partial \bm{o}_j^{(0)}}$ is an arbitrary value.

Then we discuss the value of $\frac{\partial \bm{a}_i}{\partial \bm{z}_{m}}$. For softmax function, it's easy to prove that:
\begin{equation}
\label{eq:d_softmax}
    \frac{\partial \bm{a}_s}{\partial \bm{z}_{t}} = \begin{cases}
    \bm{a}_s(1-\bm{a}_s) & ,s=t \\
    -\bm{a}_s\bm{a}_t & ,s\neq t 
    \end{cases}
\end{equation}

With \cref{eq:d_L_K,eq:d_o_a_z_k,eq:d_softmax}, we can get the final derivative value:
\begin{equation}
\begin{split}
\label{eq:d_L_K_final}
    \frac{\partial \mathcal{L}}{\partial \bm{K}_{m,n}^{(0)}} & = \sum_{j=1}^d \frac{\partial \mathcal{L}}{\partial \bm{o}_j^{(0)}}\left( \frac{\partial \bm{o}_j^{(0)}}{\partial \bm{a}_{m}^{(0)}} \frac{\partial \bm{a}_m^{(0)}}{\partial \bm{z}_{m}^{(0)}} \frac{\partial \bm{z}_m^{(0)}}{\partial \bm{K}_{m,n}^{(0)}} + \sum_{i=1,i\neq m}^l \frac{\partial \bm{o}_j^{(0)}}{\partial \bm{a}_{i}^{(0)}} \frac{\partial \bm{a}_i^{(0)}}{\partial \bm{z}_{m}^{(0)}} \frac{\partial \bm{z}_m^{(0)}}{\partial \bm{K}_{m,n}^{(0)}}\right) \\
    & = \sum_{j=1}^d \frac{\partial \mathcal{L}}{\partial \bm{o}_j^{(0)}}\left( \bm{V}_{m,j}^{(0)} \bm{a}_m^{(0)}(1-\bm{a}_m^{(0)}) \bm{q}_n + \sum_{i=1,i\neq m}^l \bm{V}_{i,j}^{(0)} (-\bm{a}_i^{(0)}\bm{a}_j^{(0)}) \bm{q}_n\right)
\end{split}
\end{equation}

With \cref{eq:K_0_V_0}, it's easy to get:
\begin{equation}
    \frac{\partial \mathcal{L}}{\partial \bm{K}_{m,n}^{(0)}} = 0
\end{equation}
thus after the first parameter update, we get
\begin{equation}
\label{eq:K_1}
    \bm{K}^{(1)} = [\bm{0}]^{l\times d}
\end{equation}

For the derivative of $\mathcal{L}$ w.r.t. $\bm{V}_{m,n}$, we have
\begin{equation}
    \frac{\partial \mathcal{L}}{\partial \bm{V}_{m,n}^{(0)}} = \frac{\partial \mathcal{L}}{\partial \bm{o}_n^{(0)}}\frac{\partial \bm{o}_n^{(0)}}{\partial \bm{V}_{m,n}^{(0)}}
\end{equation}

With \cref{eq:a_m,eq:z_m,eq:K_0_V_0}, we can get
\begin{equation}
\label{eq:d_o_v}
    \frac{\partial \bm{o}_n}{\partial \bm{V}_{m,n}^{(0)}} = \bm{a}_m^{(0)} = \frac{e^{0}}{\sum_{j=1}^l e^{0}} = \frac{1}{l}
\end{equation}
and $\frac{\partial \mathcal{L}}{\partial \bm{o}_n}$ is an arbitrary value.

So we have
\begin{equation}
\label{eq:d_L_v}
    \frac{\partial \mathcal{L}}{\partial \bm{V}_{m,n}^{(0)}} = \frac{1}{l} \frac{\partial \mathcal{L}}{\partial \bm{o}_n^{(0)}}
\end{equation}

We can find that The value of $\frac{\partial \mathcal{L}}{\partial \bm{V}_{m,n}^{(0)}}$ is independent of $m$, which means the gradients of all vectors in $\bm{V}$ are the same, formulated as
\begin{equation}
\label{eq:V_1}
    \bm{V}^{(1)} = \begin{pmatrix}
        \bm{e}^{(1)} \\
        \vdots \\
        \bm{e}^{(1)}
        \end{pmatrix} \in \mathbb{R}^{l\times d}
\end{equation}
where $\bm{e}^{(1)}$ is arbitrary vector.

(\textit{ii}) The subsequent training steps.

After the first training step, we get new parameter values according to \cref{eq:K_1,eq:V_1}. Consider the second training step. Substituting \cref{eq:K_1,eq:V_1} into \cref{eq:d_L_K_final}, we get:
\begin{equation}
\begin{split}
\label{eq:d_L_K_final_1}
    \frac{\partial \mathcal{L}}{\partial \bm{K}_{m,n}^{(1)}} & = \sum_{j=1}^d \frac{\partial \mathcal{L}}{\partial \bm{o}_j^{(1)}}\left( \bm{V}_{m,j}^{(1)} \bm{a}_m^{(1)}(1-\bm{a}_m^{(1)}) \bm{q}_n + \sum_{i=1,i\neq m}^l \bm{V}_{i,j}^{(1)} (-\bm{a}_i^{(1)}\bm{a}_j^{(1)}) \bm{q}_n\right)
\end{split}
\end{equation}

Since for all $i$, $\bm{V}_{i,j}^{(1)}$ are the same and $\bm{a}^{(1)}_i=\frac{1}{l}$, we can simplify \cref{eq:d_L_K_final_1} to
\begin{equation}
\begin{split}
\label{eq:d_L_K_final_1_v2}
    \frac{\partial \mathcal{L}}{\partial \bm{K}_{m,n}^{(1)}} & = \sum_{j=1}^d \frac{\partial \mathcal{L}}{\partial \bm{o}_j^{(1)}} \bm{V}_{m,j}^{(1)} \bm{q}_n \left(  \bm{a}_m^{(1)}(1-\bm{a}_m^{(1)})  + \sum_{i=1,i\neq m}^l (-\bm{a}_i^{(1)}\bm{a}_j^{(1)}) \right) \\
    & = \sum_{j=1}^d \frac{\partial \mathcal{L}}{\partial \bm{o}_j^{(1)}} \bm{V}_{m,j}^{(1)} \bm{q}_n \left( \frac{l-1}{l^2}  - \frac{l-1}{l^2} \right) \\
    &= 0
\end{split}
\end{equation}

So the $\bm{K}^{(2)}$ is still zero matrix:
\begin{equation}
\label{eq:K_2}
    \bm{K}^{(2)} = [\bm{0}]^{l\times d}
\end{equation}

Since $\bm{a}^{(1)} = \bm{a}^{(0)} = \frac{1}{l}$, the derivative of $\mathcal{L}$ w.r.t. $\bm{V}_{m,n}^{(1)}$ becomes
\begin{equation}
    \frac{\partial \mathcal{L}}{\partial \bm{V}_{m,n}^{(1)}} = \frac{1}{l} \frac{\partial \mathcal{L}}{\partial \bm{o}_n^{(1)}}
\end{equation}
which is still independent of $m$, thus we have
\begin{equation}
\label{eq:V_2}
    \bm{V}^{(2)} = \begin{pmatrix}
        \bm{e}^{(2)} \\
        \vdots \\
        \bm{e}^{(2)}
        \end{pmatrix} \in \mathbb{R}^{l\times d}
\end{equation}
where $\bm{e}^{(2)}$ is arbitrary vector.

It's easy to find that $\bm{K}^{(2)}, \bm{V}^{(2)}$ share the same properties as $\bm{K}^{(1)}, \bm{V}^{(1)}$, thus $\bm{K}$ remains zero throughout the subsequent training process, and $\bm{V}$ is degenerated into a matrix where all vectors are identical.

\noindent \textbf{(2)} We then discuss the scenario that key $\bm{K}$ is randomly initialized and value $\bm{V}$ is zero-initialized. It's obvious that \cref{eq:d_o_v,eq:d_L_v,eq:V_1} no longer valid, resulting in $\bm{V}^{(1)}$ becoming an arbitrary matrix. After that, all subsequent $\bm{K}^{(i)}$ and $\bm{V}^{(i)}$ can be correctly trained.

\section{Effect of Cross-task Knowledge-sharing Strategies}
\label{sec:knowledge_share}

As we discussed in the ``Limitations and future work'' section, recent literature has demonstrated the effectiveness of sharing knowledge across tasks in the class incremental learning setting, where the domain gap between tasks is relatively small. Here we implement two notable methods, G-Prompt from DualPrompt \cite{wang2022dualprompt} and Attention-based Prompting from CODA-Prompt \cite{smith2023coda}, into our DIKI framework, and test them under the challenging DCIL protocol, as shown in \cref{tab:knowledge_sharing}. G-Prompt is a set of shared prompts that are trained and utilized by all tasks, and the Attention-based Prompting mechanism weights prompts from different tasks based on key-similarity matching results, which can be naturally replaced by our distribution scores $\{S^i\}$ in Eq. ({\color{red} 9}). Results show that the integration of cross-task knowledge-sharing strategies leads to a decrease on \textit{Last} metric, while \textit{Transfer} scores remain comparable. It indicates that this degradation is caused by backward forgetting due to the sharing of task-specific knowledge. This observation underscores the need for further research into effective knowledge-sharing mechanisms specifically tailored for the DCIL setting.

\begin{table}[t]
    \setlength\tabcolsep{5pt}
    \centering
    \caption{Results of our DIKI with cross-task knowledge-sharing strategies on MTIL benchmark. The G-Prompt \cite{wang2022dualprompt} and the Attention-based Prompting (AbP) mechanism \cite{smith2023coda} are reproduced and integrated into our DIKI. Both two strategies don't work under the DCIL setting due to the severe domain gap between tasks.}
    \resizebox{0.7\columnwidth}{!}{
    \fontsize{8pt}{10pt}\selectfont
        \begin{tabular}{y{70}|x{30}x{26}x{26}|x{30}}
        \toprule
        & Transfer & Avg. & Last & Average \\
        \midrule
        \rowcolor{mygray} DIKI & \textbf{68.7} & \textbf{76.3} & \textbf{85.1} & \textbf{76.7} \\
        \quad + G-Prompt & 67.7 & 74.0 & 81.9 & 74.5 \\
        \quad + AbP & 66.5 & 72.6 & 74.3 & 71.1 \\
        \bottomrule
        \end{tabular}
    }
    \label{tab:knowledge_sharing}
\end{table}

\section{Algorithm Procedure}

We elaborate on the training and test process of our proposed DIKI in \cref{alg:train,alg:test}. We train separate $K_r^i$ and $V_r^i$ and maintain corresponding $\bm{\mu}^i$ and $\bm{\Sigma}^i$ for each task during the training phase. At test time, the $\bm{\mu}^i$ and $\bm{\Sigma}^i$ are used to identify the task information for the test sample, and  $K_r^i$ and $V_r^i$ are injected into the frozen backbone to reach better performance.

\begin{algorithm}
\caption{Training process of DIKI.}
\label{alg:train}
\textbf{Input}: Training datasets $D^i=\{x^i_j, y^i_j\}_{j=1}^{N^i}$ with class names $C^i=\{c^i_j\}_{j=1}^{N_{c}^i}$ for each task, pre-trained image encoder $f$ and text encoder $g$, learning rate $\eta$, batch size $N_{bs}$, max iterations $I_{\text{max}}$.
\begin{algorithmic}[1]
\For{$i = 1, \cdots, N$}
    \State $B_{\text{feat}}$ = \{\}
    \For{$j = 1, \cdots, N^i$}
        \State Calculate image feature $f(x^i_j)$
        \State Append $f(x^i_j)$ to $B_{\text{feat}}$
    \EndFor
    \State Calculate $\bm{\mu}^i$ and $\bm{\Sigma}^i$ with $B_{\text{feat}}$ \Comment{Eq. ({\color{red} 8})}
    \State Initialize $K_r^i$ with uniform distribution and $V_r^i$ with zero \Comment{Eq. ({\color{red} 7})}
    \For{$iter = 1, \cdots, I_{\text{max}}$}
        \State Fetch mini-batch samples $\{x^i_j, y^i_j\}_{j=1}^{N_{bs}}$ from $D^i$
        \State Insert $K_r^i$ and $V_r^i$ to $f$ and $g$, get $f'$ and $g'$ \Comment{Eq. ({\color{red} 6})}
        \State Calculate image features $\{f'(x^i_j)\}_{j=1}^{N_{bs}}$
        \State Calculate class name text embeddings $\{g'(c^i_j)\}_{j=1}^{N_{c}^i}$
        \State Compute cosine similarities between them $s_{j,k} = \Braket{f'(x^i_j), g'(c^i_k)}$
        \State Get final predictions with softmax $p_{j,k} = \frac{\exp(s_{j,k})}{\sum_k'\exp(s_{j,k'})}$
        \State Calculate Cross-Entropy loss $\mathcal{L}=\text{CELoss}(p, y^i)$ 
        \State Update $K_r^i = K_r^i-\eta\nabla_{K_r^i}\mathcal{L}$
        \State Update $V_r^i = V_r^i-\eta\nabla_{V_r^i}\mathcal{L}$
    \EndFor
\EndFor
\end{algorithmic}
\end{algorithm}

\begin{algorithm}
\caption{Test process of DIKI.}
\label{alg:test}
\textbf{Input}: Test dataset $D_t=\{x_j\}_{j=1}^{N_{t}}$ with class names $C^i=\{c^i_j\}_{j=1}^{N_{c}^i}$, pre-trained image encoder $f$ and text encoder $g$, currently trained parameters $\{K_r^i, V_r^i\}_{i=1}^{N_{\text{cur}}}$, distribution parameters $\{\bm{\mu}^i, \bm{\Sigma}^i\}_{i=1}^{N_{\text{cur}}}$.
\begin{algorithmic}[1]
\For{$x$ in $D_t$}
    \State Calculate image feature $f(x)$
    \State Compute the logarithm of the probability density $\{S^i\}_{i=1}^{N_{\text{cur}}}$ \Comment{Eq. ({\color{red} 9})}
    \State Get max value $\hat{S}$ and corresponding index $s$
    \State Calculate the integration calibration weight $\mathcal{M}(\hat{S})$ \Comment{Eq. ({\color{red} 10})}
    \State Insert $K_r^s$ and $V_r^s$ to $f$ and $g$, get $f'$ and $g'$ \Comment{Eq. ({\color{red} 6})}
    \State Calculate image feature $f'(x)$ with calibration weight \Comment{Eq. ({\color{red} 10})}
    \State Calculate text embeddings $\{g'(c^s_j)\}_{j=1}^{N_{c}^s}$ with calibration weight \Comment{Eq. ({\color{red} 10})}
    \State Compute cosine similarities between them $s_j = \Braket{f'(x), g'(c^s_j)}$
    \State Compute predictions with softmax $p_j = \frac{\exp(s_j)}{\sum_k\exp(s_k)}$ and get  classification results
\EndFor
\end{algorithmic}
\end{algorithm}

\section{Details about Reproduction}

\textbf{L2P}~\cite{wang2022learning} We reproduce L2P on CLIP by simply prompting both the visual and text encoders. In the original L2P paper, the updated prompts are selected by a key-matching mechanism during the training stage, and the diversity of prompt-selection is guaranteed by a frequency-based weight technique. However, in their official code repository\footnote{\url{https://github.com/google-research/l2p}}, they mask specific prompts for different tasks. We follow the implementation of their official code.

At test time, we only select the top-1 prompt because, under the domain-class incremental learning setting, it is challenging to extract domain-invariant knowledge, and most of the learned knowledge is non-shareable. Adopting the original setting (i.e., top-5) would significantly degrade performance. 

Regarding other hyper-parameters, the prompt length is set at 32, the learning rate is set to 0.05, and the weight of the key match loss is set at 5. The remaining training settings are the same as those in our DIKI.

\noindent\textbf{DualPrompt}~\cite{wang2022dualprompt} Similar to L2P, we simply prompt both the visual and text encoders to adapt DualPrompt to CLIP. Following the original paper, the prefix tuning is applied. DualPrompt separate prompts into G(eneral)-Prompt and E(xpert)-Prompt. However, similar to our discussion in \cref{sec:knowledge_share}, we find that the G-Prompt will cause a significant performance drop. This is because the knowledge learned in different tasks is mostly non-shareable, different from class-incremental settings. So we remove the G-Prompt to prevent degradation.

Regarding other hyper-parameters, the prompt length and depth are both set at 8, the learning rate is set to 5, and the weight of the key match loss is set at 0.1. The remaining training settings are the same as those in our DIKI.

\noindent\textbf{S-Prompts}~\cite{wang2022s} S-Prompts is originally proposed on CLIP model, we simply adopt it on MTIL benchmark. Toward hyper-parameters, the prompt length is set at 32, the learning rate is set to 0.05. The remaining training settings are the same as those in our DIKI.

\section{Details about MTIL Benchmark}

\subsection{Datasets}

Authors introduced two different dataset orders in the original paper \cite{wang2022learning}. The first, Order-I, follows an alphabetical sequence: Aircraft, Caltech101, CIFAR100, DTD, EuroSAT, Flowers, Food, MNIST, OxfordPet, StanfordCars, SUN397. The second, Order-II, is arranged randomly: StanfordCars, Food, MNIST, OxfordPet, Flowers, SUN397, Aircraft, Caltech101, DTD, EuroSAT, CIFAR100. Order-I is adopted for results presented in Tab. 1 of the manuscript, and experiments were also conducted on Order-II, as indicated in \cref{tab:complete_res_ours_order2,tab:complete_res_l2p_order2,tab:complete_res_dualprompt_order2,tab:complete_res_sprompts_order2}.

For our modified MTIL-FS benchmark in a few-shot setting, we only use 16 samples per class for model training. We exclude EuroSAT, MNIST, and OxfordPet due to their severely insufficient training samples caused by their small number of classes. More discussion on this can be found in \cref{sec:more_limitation}. To maintain reproducibility, we adopt the data splits from the official repository of CoOp \cite{zhou2022learning}, which is widely used by many CLIP-based few-shot learning works. Since CIFAR100 is not included by CoOp, we generate its training set by random selection with a random seed 42.

\subsection{Metrics}

Here we formulate the \textit{Transfer}, \textit{Avg.} and \textit{Last} metrics.

Assume that $p_{j}^{(i)}$ is the model's accuracy on task $j$ after being trained on task $i$, then the Transfer, Avg and Last metrics for task $j$ can be calculated as:

\begin{equation}
\begin{gathered}
    \text{Transfer}_j = \frac{1}{j-1} \sum_{i=1}^{j-1} p_{j}^{(i)}, \quad j=2,3,\cdots,N \\
    \text{Avg}_j = \frac{1}{N} \sum_{i=1}^{N} p_{j}^{(i)}, \quad j=1,2,\cdots,N \\
    \text{Last}_j = p_{j}^{(N)}, \quad j=1,2,\cdots,N
\end{gathered}
\end{equation}
where $N$ is the number of tasks. It's clear that \textit{Transfer} metric can indicate the zero-shot capability while \textit{Last} metric shows the extent of backward forgetting.

\begin{figure}[t]
    \centering
    \includegraphics[width=0.5\linewidth]{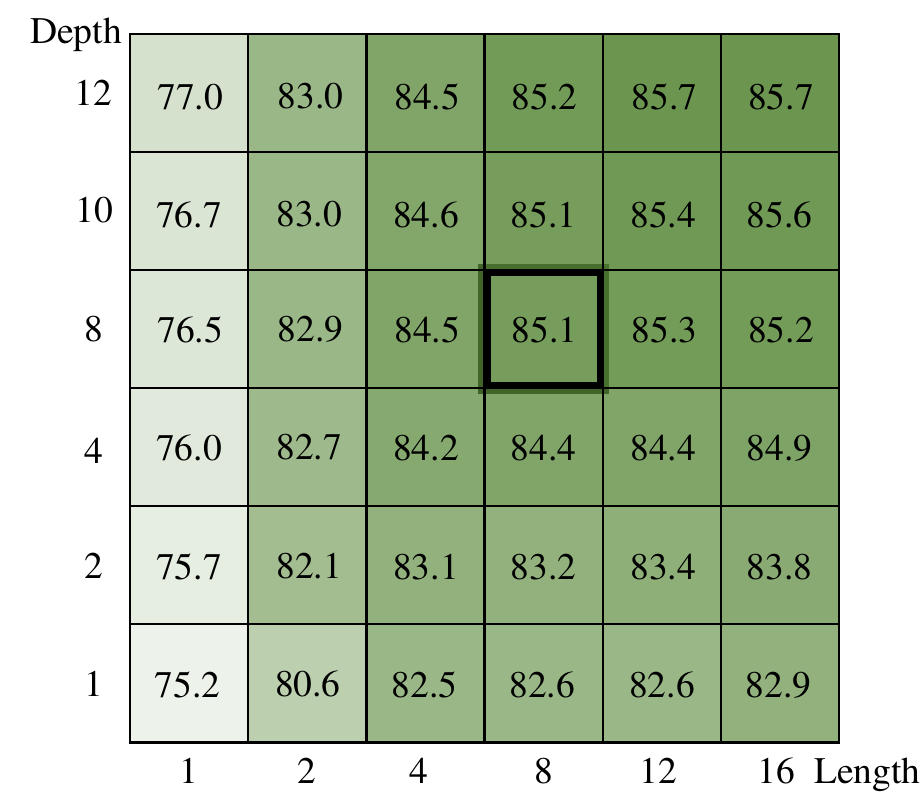} 
    \caption{\textit{Last} score (\%) with different IKI structure hyper-parameters.  Setting high-lighted in bold was chosen in our all experiments.}
    \label{fig:para_tuning}
\end{figure}

\section{Additional Results}

\cref{tab:finalacc_order2} shows the results of \textit{Transfer}, \textit{Avg.}, and \textit{Last} metrics on MTIL benchmark with Order-II, and \cref{tab:fewshot_full} provides full results of different continue learning methods on our modified 16-shot MTIL-FS benchmark. Our DIKI shows consistent improvements compared to previous methods.

For the selection of hyper-parameters, we perform a search on the structure parameters of our introduced $K_r$ and $V_r$ in our IKI. Length denotes the vector number $l$ in Eq. ({\color{red} 6}), and depth indicates the number of layers implemented, starting from the input layer. Given that our distribution-aware attention scaling scheme ensures minimal variation in the \textit{Transfer} metric across different hyper-parameters, we focus on demonstrating the Last scores with varying parameters, as depicted in \cref{fig:para_tuning}. Generally, an increase in the number of trainable parameters correlates with improved model accuracy. But we observe diminishing returns when depth and length exceed 8, thus we select a configuration of (8, 8) for all our experiments.

We also record the task assignment results during the test phase, as shown in \cref{tab:assignment_accuracy}. When the model is only trained on task $i$ and earlier tasks, the task assignment results for samples from unseen tasks $i+1, \cdots, N$ are always incorrect. Thus we omit the meaningless upper triangular area and only consider the rest part. Results demonstrate that our task assignment on learned tasks holds high accuracy. Note that the misassignment of samples from unseen tasks is also resolved by our distribution-aware integration calibration.

Additionally, \cref{tab:complete_res_ours,tab:complete_res_l2p,tab:complete_res_dualprompt,tab:complete_res_sprompts} shows per training step accuracies of different methods on MTIL benchmark with Order-I, \cref{tab:complete_res_ours_order2,tab:complete_res_l2p_order2,tab:complete_res_dualprompt_order2,tab:complete_res_sprompts_order2} shows that results with Order-II, and \cref{tab:complete_res_ours_fs,tab:complete_res_l2p_fs,tab:complete_res_dualprompt_fs,tab:complete_res_sprompts_fs,tab:complete_res_zscl_fs} shows per training step accuracies of different methods on MTIL-FS benchmark.

\section{More Limitations and Future Directions}
\label{sec:more_limitation}

Because of the use of parameter-efficient fine-tuning techniques, we could achieve high performance with significantly fewer trainable parameters. However, the knowledge learned by such a small number of parameters is definitely less than that obtained through full-parameter fine-tuning, as evident from the per-step accuracies table. For example, if we compare \cref{tab:complete_res_ours} and the Tab. 11 from the ZSCL paper \cite{zheng2023preventing}, it's easy to find that our DIKI achieves lower accuracy when the model is tested immediately after training on the some datasets compared to ZSCL. The reason for our higher final performance is that DIKI can precisely memorize previously trained knowledge, while ZSCL suffers from backward forgetting issues. One future direction is to find a parameter-efficient fine-tuning method that can store more information to mitigate the gap.

In our modified MTIL-FS benchmark, we exclude three datasets for their lack of classes. This is because task-specific prompt learning methods (L2P \cite{wang2022learning}, DualPrompt \cite{wang2022dualprompt}, S-Prompts \cite{wang2022s}, DIKI) can't obtain robust task identities with such limited training samples, leading to test performance degradation due to the inaccurate task assignments. This indicates a prevalent challenge associated with task-specific prompt learning methods: their heavy dependence on accurate task assignments. ZSCL \cite{zheng2023preventing} leverages knowledge distillation from large-scale reference datasets to alleviate the need for the task assignment process, which requires extensive computation and storage resources. Future works can tackle this issue by developing more robust task identification techniques or introducing task assignment-free prompt learning methods.

\begin{table*}[t]
\setlength\tabcolsep{5pt}
\centering
\caption{\textit{Transfer}, \textit{Avg.}, and \textit{Last} scores (\%) of different continue learning methods on MTIL benchmark with Order-II. Metric ``transfer'' represents the model zero-shot ability retention after being trained on each task. $\dag$ means we reproduce the original methods on vision-language models. 
}
\label{tab:finalacc_order2}
{
\fontsize{8pt}{10pt}\selectfont
\resizebox{1\textwidth}{!}{
\begin{tabular}{y{70}x{8}x{25}|*{11}{x{17}}|x{22}}
\toprule
 & \rot{Extra data} & \rot{\# Param.} & \rot{Cars} & \rot{Food} & \rot{MNIST} & \rot{OxfordPet} & \rot{Flowers} & \rot{SUN397} & \rot{Aircraft} & \rot{Caltech101} & \rot{DTD} & \rot{EuroSAT} & \rot{CIFAR100} & \rot{Average} \\ \midrule

\quad Zero-shot & & & 65.8 & 85.8 & 59.5 & 89.1 & 71.4 & 62.6 & 24.8 & 92.9 & 43.8 & 47.7 & 68.4 & 64.7 \\
\quad Upper Bound & & & 89.6 & 92.7 & 99.6 & 94.7 & 97.5 & 81.8 & 62.0 & 96.2 & 79.5 & 98.9 & 89.6 & 89.3 \\ \midrule
\midrule

\textbf{Transfer} \\
\quad LwF~\cite{li2017learning} & $\checkmark$ & 211 M & & 87.8 & 58.5 & 71.9 & 46.6 & 57.3 & 12.8 & 81.4 & 34.5 & 34.5 & 46.8 & 53.2 \\
\quad iCaRL~\cite{rebuffi2017icarl} & $\checkmark$ & 211 M & & 86.1 & 51.8 & 67.6 & 50.4 & 57.9 & 11.0 & 72.3 & 31.2 & 32.7 & 48.1 & 50.9 \\
\quad LwF-VR~\cite{ding2022don} & $\checkmark$ & 211 M & & 88.2 & 57.0 & 71.4 & 50.0 & 58.0 & 13.0 & 82.0 & 34.4 & 29.3 & 47.6 & 53.1 \\
\quad WiSE-FT~\cite{wortsman2022robust} & $\checkmark$ & 211 M & & 87.2 & 57.6 & 67.0 & 45.0 & 54.0 & 12.9 & 78.6 & 35.5 & 28.4 & 44.3 & 51.0  \\
\quad ZSCL$^*$~\cite{zheng2023preventing} & $\checkmark$ & 211 M & & \textbf{88.8} & 56.7 & 75.5 & 58.8 & 62.5 & 16.1 & 87.0 & 42.0 & 44.0 & 66.5 & 59.8 \\
\quad ZSCL~\cite{zheng2023preventing} & $\checkmark$ & 211 M & & 88.3 & 57.5 & 84.7 & 68.1 & \textbf{64.8} & 21.1 & 88.2 & \textbf{45.3} & \textbf{55.2} & \textbf{68.2} & 64.2 \\ \midrule
\quad L2P$^\dag$~\cite{wang2022learning} & $\times$ & 0.5 M & & 70.6 & 30.7 & 78.3 & 42.8 & 38.3 & 17.4 & 75.3 & 27.4 & 23.1 & 20.7 & 42.5 \\
\quad DualPmt.$^\dag$~\cite{wang2022dualprompt} & $\times$ & 1.8 M & & 79.9 & 46.9 & 85.2 & 51.3 & 45.1 & 9.3 & 82.7 & 29.9 & 42.9 & 47.2 & 52.1 \\
\quad S-Prompts~\cite{wang2022s} & $\times$ & 0.5 M & & 59.8 & 46.2 & 67.7 & 47.5 & 43.8 & 13.5 & 76.8 & 31.4 & 22.6 & 43.5 & 45.3 \\
\rowcolor{mygray} \quad DIKI & $\times$ & 1.8 M & & 85.8 & \textbf{59.8} & \textbf{89.1} & \textbf{71.8} & 62.6 & \textbf{24.3} & \textbf{93.3} & 42.7 & 46.8 & 67.8 & \textbf{64.4} \\ \midrule
\midrule

\textbf{Avg.} \\
\quad LwF~\cite{li2017learning} & $\checkmark$ & 211 M & 49.0 & 77.0 & \textbf{92.1} & 85.9 & 66.5 & 67.2 & 20.9 & 84.7 & 44.6 & 45.5 & 50.5 & 62.2 \\
\quad iCaRL~\cite{rebuffi2017icarl} & $\checkmark$ & 211 M & 52.0 & 75.9 & 77.4 & 74.6 & 58.4 & 59.3 & 11.7 & 79.6 & 42.1 & 43.2 & 51.7 & 56.9 \\
\quad LwF-VR~\cite{ding2022don} & $\checkmark$ & 211 M & 44.9 & 75.8 & 91.8 & 85.3 & 63.5 & 67.6 & 16.9 & 84.9 & 44.0 & 40.6 & 51.3 & 60.6 \\
\quad WiSE-FT~\cite{wortsman2022robust} & $\checkmark$ & 211 M & 52.6 & 79.3 & 91.9 & 83.9 & 63.4 & 65.2 & 23.3 & 83.7 & 45.4 & 40.0 & 48.2 & 61.5 \\
\quad ZSCL$^*$~\cite{zheng2023preventing} & $\checkmark$ & 211 M & 72.0 & 89.8 & 91.7 & 87.9 & 78.8 & 71.5 & \textbf{35.1} & 89.0 & 51.4 & 53.9 & 68.5 & 71.8 \\
\quad ZSCL~\cite{zheng2023preventing} & $\checkmark$ & 211 M & 81.7 & \textbf{91.3} & 91.1 & 91.0 & 82.9 & \textbf{72.5} & 33.6 & 89.7 & \textbf{53.3} & \textbf{62.8} & \textbf{69.9} & \textbf{74.5} \\ \midrule
\quad L2P$^\dag$~\cite{wang2022learning} & $\times$ & 0.5 M &80.1 & 87.4 & 86.7 & 89.6 & 76.8 & 59.1 & 27.7 & 79.5 & 39.9 & 34.6 & 26.5 & 62.5 \\
\quad DualPmt.$^\dag$~\cite{wang2022dualprompt} & $\times$ & 1.8 M & 78.6 & 88.4 & 89.7 & 91.7 & 80.0 & 62.4 & 23.2 & 85.0 & 41.3 & 51.6 & 50.7 & 67.5 \\
\quad S-Prompts~\cite{wang2022s} & $\times$ & 0.5 M & 79.2 & 86.5 & 89.5 & 87.0 & 78.2 & 61.5 & 25.5 & 83.6 & 41.9 & 36.3 & 47.2 & 65.1 \\
\rowcolor{mygray} \quad DIKI & $\times$ & 1.8 M & \textbf{81.9} & 88.9 & \textbf{92.1} & \textbf{92.8} & \textbf{87.7} & 70.3 & 34.3 & \textbf{94.2} & 51.5 & 56.1 & 69.5 & \textbf{74.5} \\ \midrule
\midrule

\textbf{Last} \\
\quad LwF~\cite{li2017learning} & $\checkmark$ & 211 M & 34.6 & 69.6 & 99.3 & 88.7 & 61.1 & 72.5 & 32.5 & 88.1 & 65.6 & 90.9 & 87.9 & 71.9 \\
\quad iCaRL~\cite{rebuffi2017icarl} & $\checkmark$ & 211 M & 46.0 & 81.5 & 91.3 & 82.8 & 66.5 & 72.2 & 16.3 & 91.6 & 68.1 & 83.2 & 87.8 & 71.6 \\
\quad LwF-VR~\cite{ding2022don} & $\checkmark$ & 211 M & 27.4 & 61.2 & 99.4 & 86.3 & 60.6 & 70.7 & 23.4 & 88.0 & 61.3 & 84.3 & 88.1 & 68.3 \\
\quad WiSE-FT~\cite{wortsman2022robust} & $\checkmark$ & 211 M & 35.6 & 76.9 & \textbf{99.5} & 89.1 & 62.1 & 71.8 & 27.8 & 90.8 & 67.0 & 85.6 & 87.6 & 72.2 \\
\quad ZSCL$^*$~\cite{wang2022learning} & $\checkmark$ & 211 M & 63.5 & 89.6 & 99.2 & 92.4 & 84.5 & \textbf{78.3} & \textbf{55.2} & 92.4 & 74.6 & 97.4 & \textbf{88.6} & 83.3 \\
\quad ZSCL~\cite{wang2022learning} & $\checkmark$ & 211 M & 78.2 & \textbf{91.1} & 97.6 & 92.5 & 87.4 & 78.2 & 45.0 & 92.3 & 72.7 & 96.2 & 86.3 & 83.4 \\
\midrule
\quad L2P$^\dag$~\cite{wang2022learning} & $\times$ & 0.5 M & 80.1 & 89.1 & 99.1 & 93.8 & 96.2 & 76.5 & 40.1 & 86.9 & 73.5 & 86.3 & 84.2 & 82.3 \\
\quad DualPmt.$^\dag$~\cite{wang2022dualprompt} & $\times$ & 1.8 M & 78.6 & 89.3 & 99.2 & 94.1 & 96.5 & 76.8 & 39.8 & 89.0 & 71.6 & 90.7 & 84.9 & 82.8 \\
\quad S-Prompts~\cite{wang2022s} & $\times$ & 0.5 M & 79.2 & 89.1 & 99.1 & \textbf{94.3} & 95.8 & 76.3 & 39.9 & 95.5 & 70.1 & 97.6 & 84.4 & 83.8 \\
\rowcolor{mygray} \quad DIKI & $\times$ & 1.8 M & \textbf{81.9} & 89.2 & 99.4 & \textbf{94.3} & \textbf{96.8} & 76.7 & 46.3 & \textbf{95.9} & \textbf{74.8} & \textbf{98.3} & 86.6 & \textbf{85.5} \\ 
\bottomrule
\end{tabular}}
}
\end{table*}

\begin{table*}[t]
\setlength\tabcolsep{6pt}
\centering
\caption{Full results of different continue learning methods on 16-shot MTIL-FS benchmark. $\dag$ means we reproduce the original methods on vision-language models.}
\label{tab:fewshot_full}
\fontsize{8pt}{10pt}\selectfont
\resizebox{\textwidth}{!}{
\begin{tabular}{y{70}x{8}x{25}|*{8}{x{17}}|x{22}}
\toprule
 & \rot{Extra data} & \rot{\# Param.} & \rot{Aircraft} & \rot{Caltech101} & \rot{CIFAR100} & \rot{DTD} & \rot{Flowers} & \rot{Food} & \rot{Cars} & \rot{SUN397} & \rot{Average} \\ \midrule

\quad Zero-shot & & & 24.8 & 92.9 & 68.4 & 43.8 & 71.4 & 85.8 & 65.8 & 62.6 & 64.4 \\
\quad Upper Bound & & & 62.0 & 96.2 & 89.6 & 79.5 & 97.5 & 92.7 & 89.6 & 81.8 & 86.1 \\ \midrule
\midrule

\textbf{Transfer} \\
\quad ZSCL~\cite{zheng2023preventing} & $\checkmark$ & 211 M & & 87.3 & 67.7 & \textbf{45.4} & 67.8 & \textbf{86.6} & 59.7 & 63.4 & 68.3 \\ \midrule
\quad L2P$^\dag$~\cite{wang2022learning} & $\times$ & 0.5 M & & 66.7 & 54.3 & 30.6 & 47.3 & 71.5 & 54.6 & 52.4 & 53.9 \\
\quad DualPmt.$^\dag$~\cite{wang2022dualprompt} & $\times$ & 1.8 M & & 78.8 & 64.4 & 32.0 & 51.7 & 77.5 & 49.4 & 51.3 & 57.9 \\
\quad S-Prompts~\cite{wang2022s} & $\times$ & 0.5 M & & 70.3 & 52.7 & 31.5 & 54.8 & 74.0 & 55.4 & 50.0 & 55.5 \\
\quad DIKI & $\times$ & 1.8 M & & \textbf{92.7} & \textbf{68.8} & 44.1 & \textbf{70.0} & 86.2 & \textbf{65.1} & \textbf{65.5} & \textbf{70.3} \\ \midrule
\midrule

\textbf{Avg.} \\
\quad ZSCL~\cite{zheng2023preventing} & $\checkmark$ & 211 M & 33.5 & 90.5 & 74.7 & \textbf{58.5} & 79.7 & \textbf{87.7} & 64.8 & 64.8 & 69.3 \\ \midrule
\quad L2P$^\dag$~\cite{wang2022learning} & $\times$ & 0.5 M & 30.2 & 84.5 & 70.1 & 51.9 & 69.6 & 77.1 & 60.0 & 55.2 & 62.3 \\
\quad DualPmt.$^\dag$~\cite{wang2022dualprompt} & $\times$ & 1.8 M & 36.5 & 89.5 & 72.5 & 52.7 & 72.3 & 80.8 & 56.1 & 54.2 & 64.3 \\
\quad S-Prompts~\cite{wang2022s} & $\times$ & 0.5 M & 30.6 & 86.8 & 70.0 & 51.7 & 74.3 & 78.5 & 60.7 & 53.0 & 63.2 \\
\quad DIKI & $\times$ & 1.8 M & \textbf{41.3} & \textbf{95.3} & \textbf{76.5} & \textbf{58.5} & \textbf{82.2} & 86.4 & \textbf{68.2} & \textbf{66.6} & \textbf{71.9} \\ \midrule
\midrule

\textbf{Last} \\
\quad ZSCL~\cite{wang2022learning} & $\checkmark$ & 211 M & 27.7 & 90.9 & 74.4 & 64.7 & 90.2 & \textbf{89.2} & \textbf{80.6} & 74.6 & 74.0 \\
\midrule
\quad L2P$^\dag$~\cite{wang2022learning} & $\times$ & 0.5 M & 30.2 & 87.1 & 75.4 & 64.7 & 91.9 & 86.4 & 76.1 & \textbf{74.7} & 73.3 \\
\quad DualPmt.$^\dag$~\cite{wang2022dualprompt} & $\times$ & 1.8 M & 36.5 & 91.0 & 75.1 & 65.1 & 92.9 & 86.2 & 76.2 & 74.2 & 74.7 \\
\quad S-Prompts~\cite{wang2022s} & $\times$ & 0.5 M & 30.6 & 89.2 & 75.8 & 63.8 & 93.9 & 86.2 & 76.7 & 73.9 & 73.8 \\
\quad DIKI & $\times$ & 1.8 M & \textbf{41.3} & \textbf{95.6} & \textbf{79.0} & \textbf{67.3} & \textbf{94.4} & 86.8 & 77.6 & 74.4 & \textbf{77.1} \\ 
\bottomrule
\end{tabular}
}
\end{table*}

\begin{table*}[t]
\setlength\tabcolsep{5pt}
\centering
\caption{Task assignment accuracy (\%) for test data. Each row represents the assignment accuracy on every dataset of the model trained after the corresponding task.}
\label{tab:assignment_accuracy}
\fontsize{8pt}{10pt}\selectfont
\resizebox{0.9\textwidth}{!}{
\begin{tabular}{@{}lccccccccccc@{}}
\toprule
& \rot{Aircraft} & \rot{Caltech101} & \rot{CIFAR100} & \rot{DTD} & \rot{EuroSAT} & \rot{Flowers} & \rot{Food} & \rot{MNIST} & \rot{OxfordPet} & \rot{Cars} & \rot{SUN397} \\ \midrule
Aircraft & 100.0 & \textbf{-} & \textbf{-} & \textbf{-} & \textbf{-} & \textbf{-} & \textbf{-} & \textbf{-} & \textbf{-} & \textbf{-} & \textbf{-} \\
Caltech101 & 99.0 & 99.8 & \textbf{-} & \textbf{-} & \textbf{-} & \textbf{-} & \textbf{-} & \textbf{-} & \textbf{-} & \textbf{-} & \textbf{-} \\
CIFAR100 & 99.0 & 99.8 & 99.6 & \textbf{-} & \textbf{-} & \textbf{-} & \textbf{-} & \textbf{-} & \textbf{-} & \textbf{-} & \textbf{-} \\
DTD & 99.0 & 99.6 & 99.6 & 97.5 & \textbf{-} & \textbf{-} & \textbf{-} & \textbf{-} & \textbf{-} & \textbf{-} & \textbf{-} \\
EuroSAT & 99.0 & 99.6 & 99.6 & 97.5 & 99.4 & \textbf{-} & \textbf{-} & \textbf{-} & \textbf{-} & \textbf{-} & \textbf{-} \\
Flowers & 99.0 & 99.1 & 99.6 & 97.0 & 99.4 & 97.7 & \textbf{-} & \textbf{-} & \textbf{-} & \textbf{-} & \textbf{-} \\
Food & 99.0 & 98.9 & 99.6 & 95.9 & 99.4 & 97.7 & 99.6 & \textbf{-} & \textbf{-} & \textbf{-} & \textbf{-} \\
MNIST & 99.0 & 98.9 & 99.6 & 95.9 & 99.4 & 97.7 & 99.6 & 99.6 & \textbf{-} & \textbf{-} & \textbf{-} \\
OxfordPet & 99.0 & 98.4 & 99.6 & 95.9 & 99.4 & 97.7 & 99.6 & 99.6 & 96.8 & \textbf{-} & \textbf{-} \\
Cars & 99.0 & 98.3 & 99.6 & 95.9 & 99.4 & 97.7 & 99.6 & 99.6 & 96.8 & 99.7 & \textbf{-} \\
SUN397 & 98.3 & 95.5 & 99.5 & 94.3 & 99.3 & 97.7 & 99.0 & 99.6 & 96.0 & 99.1 & 99.3 \\
\bottomrule
\end{tabular}%
}
\end{table*}

\begin{table*}[t]
\setlength\tabcolsep{5pt}
\centering
\caption{Accuracy (\%) of our DIKI on the MTIL benchmark with order-I. Each row represents the performance on every dataset of the model trained after the corresponding task. \textcolor{transfer}{Transfer}, \textcolor{avg}{Avg.}, and \textcolor{last}{Last} metrics are shown in color.}
\label{tab:complete_res_ours}
\fontsize{8pt}{10pt}\selectfont
\resizebox{\textwidth}{!}{
\begin{tabular}{@{}lcccccccccccc@{}}
\toprule
& \rot{Aircraft} & \rot{Caltech101} & \rot{CIFAR100} & \rot{DTD} & \rot{EuroSAT} & \rot{Flowers} & \rot{Food} & \rot{MNIST} & \rot{OxfordPet} & \rot{Cars} & \rot{SUN397} \\ \midrule
Transfer & & 92.9 & 69.0 & 43.2 & 48.2 & 67.4 & 85.2 & 63.0 & 87.9 & 63.8 & 66.2 & \colorbox{transfer}{68.7} \\ \midrule
Aircraft & 45.2 & 92.9 & 68.4 & 43.9 & 47.7 & 71.3 & 85.8 & 59.8 & 89.2 & 65.8 & 62.4 \\
Caltech101 & 45.1 & 95.7 & 69.5 & 42.9 & 49.0 & 66.4 & 85.8 & 50.3 & 87.7 & 63.5 & 66.7 \\
CIFAR100 & 45.1 & 95.7 & 86.3 & 42.9 & 47.4 & 66.4 & 85.8 & 66.1 & 87.7 & 63.5 & 66.7 \\
DTD & 45.1 & 95.7 & 86.3 & 72.9 & 48.7 & 66.3 & 84.5 & 66.1 & 87.7 & 63.5 & 66.6 \\
EuroSAT & 45.1 & 95.7 & 86.3 & 72.9 & 98.0 & 66.3 & 84.5 & 66.1 & 87.7 & 63.5 & 66.6 \\
Flowers & 45.1 & 95.7 & 86.3 & 72.9 & 98.0 & 97.0 & 84.5 & 66.1 & 87.7 & 63.5 & 66.6 \\
Food & 45.1 & 95.7 & 86.3 & 72.9 & 98.0 & 97.0 & 89.2 & 66.1 & 87.7 & 63.5 & 66.6 \\
MNIST & 45.1 & 95.7 & 86.3 & 72.9 & 98.0 & 97.0 & 89.2 & 99.4 & 87.7 & 63.5 & 66.6 \\
OxfordPet & 45.1 & 95.8 & 86.3 & 72.9 & 98.0 & 97.0 & 89.2 & 99.4 & 94.2 & 63.5 & 66.6 \\
Cars & 45.1 & 95.8 & 86.3 & 72.9 & 98.0 & 97.0 & 89.2 & 99.4 & 94.2 & 81.5 & 66.6 \\
SUN397 & 45.2 & 95.7 & 86.3 & 72.9 & 98.0 & 97.0 & 89.2 & 99.4 & 94.2 & 81.6 & 76.6 & \colorbox{last}{85.1} \\ \midrule
Avg. & 45.1 & 95.5 & 83.1 & 64.8 & 79.9 & 83.5 & 87.0 & 76.2 & 89.6 & 67.0 & 67.1 & \colorbox{avg}{76.3} \\
\bottomrule
\end{tabular}%
}
\end{table*}

\begin{table*}[t]
\setlength\tabcolsep{5pt}
\centering
\caption{Accuracy (\%) of L2P on the MTIL benchmark with order-I. Each row represents the performance on every dataset of the model trained after the corresponding task. \textcolor{transfer}{Transfer}, \textcolor{avg}{Avg.}, and \textcolor{last}{Last} metrics are shown in color.}
\label{tab:complete_res_l2p}
\fontsize{8pt}{10pt}\selectfont
\resizebox{\textwidth}{!}{
\begin{tabular}{@{}lcccccccccccc@{}}
\toprule
& \rot{Aircraft} & \rot{Caltech101} & \rot{CIFAR100} & \rot{DTD} & \rot{EuroSAT} & \rot{Flowers} & \rot{Food} & \rot{MNIST} & \rot{OxfordPet} & \rot{Cars} & \rot{SUN397} \\ \midrule
Transfer & & 65.6 & 50.9 & 30.4 & 41.4 & 49.3 & 71.8 & 36.3 & 77.5 & 55.3 & 53.4 & \colorbox{transfer}{53.2} \\ \midrule
Aircraft & 38.0 & 65.6 & 40.7 & 16.6 & 26.4 & 22.1 & 43.9 & 39.9 & 54.8 & 57.8 & 41.8 \\
Caltech101 & 38.0 & 87.1 & 61.1 & 37.3 & 43.7 & 56.4 & 77.1 & 47.4 & 80.7 & 55.0 & 54.2 \\
CIFAR100 & 38.0 & 87.1 & 84.2 & 37.3 & 47.6 & 56.4 & 77.1 & 33.4 & 80.7 & 55.0 & 54.2 \\
DTD & 38.0 & 87.1 & 84.2 & 72.9 & 47.6 & 55.8 & 77.6 & 33.4 & 80.7 & 55.0 & 54.9 \\
EuroSAT & 38.0 & 87.1 & 84.2 & 72.9 & 97.4 & 55.8 & 77.6 & 33.4 & 80.7 & 55.0 & 54.7 \\
Flowers & 38.0 & 87.1 & 84.2 & 72.9 & 97.4 & 96.1 & 77.6 & 33.4 & 80.7 & 55.0 & 54.6 \\
Food & 38.0 & 87.1 & 84.2 & 72.9 & 97.4 & 96.1 & 89.2 & 33.4 & 80.7 & 55.0 & 54.6 \\
MNIST & 38.0 & 87.1 & 84.2 & 72.9 & 86.0 & 96.1 & 89.2 & 99.0 & 80.7 & 55.0 & 54.6 \\
OxfordPet & 38.0 & 87.1 & 84.2 & 72.9 & 86.0 & 96.1 & 89.2 & 99.0 & 94.1 & 55.0 & 54.9 \\
Cars & 38.0 & 87.1 & 84.2 & 72.9 & 86.0 & 96.1 & 89.2 & 99.0 & 94.1 & 79.6 & 54.9 \\
SUN397 & 38.0 & 87.1 & 84.2 & 72.9 & 86.0 & 96.1 & 89.2 & 99.0 & 94.1 & 79.6 & 76.0 & \colorbox{last}{82.0} \\ \midrule
Avg. & 38.0 & 85.2 & 78.2 & 61.3 & 72.9 & 74.9 & 79.7 & 59.1 & 82.0 & 59.7 & 55.4 & \colorbox{avg}{67.9} \\
\bottomrule
\end{tabular}%
}
\end{table*}

\begin{table*}[t]
\setlength\tabcolsep{5pt}
\centering
\caption{Accuracy (\%) of DualPrompt on the MTIL benchmark with order-I. Each row represents the performance on every dataset of the model trained after the corresponding task. \textcolor{transfer}{Transfer}, \textcolor{avg}{Avg.}, and \textcolor{last}{Last} metrics are shown in color.}
\label{tab:complete_res_dualprompt}
\fontsize{8pt}{10pt}\selectfont
\resizebox{\textwidth}{!}{
\begin{tabular}{@{}lcccccccccccc@{}}
\toprule
& \rot{Aircraft} & \rot{Caltech101} & \rot{CIFAR100} & \rot{DTD} & \rot{EuroSAT} & \rot{Flowers} & \rot{Food} & \rot{MNIST} & \rot{OxfordPet} & \rot{Cars} & \rot{SUN397} \\ \midrule
Transfer & & 56.7 & 51.4 & 28.7 & 33.7 & 45.6 & 70.9 & 59.5 & 77.7 & 49.5 & 50.4 & \colorbox{transfer}{52.4} \\ \midrule
Aircraft & 37.8 & 56.7 & 35.8 & 18.5 & 30.6 & 31.6 & 52.5 & 45.0 & 61.6 & 46.7 & 20.1 \\
Caltech101 & 37.8 & 87.1 & 67.0 & 33.9 & 53.7 & 52.9 & 73.7 & 48.0 & 80.0 & 49.9 & 54.4 \\
CIFAR100 & 37.8 & 87.1 & 84.6 & 33.9 & 25.2 & 52.9 & 73.7 & 64.7 & 80.0 & 49.9 & 54.4 \\
DTD & 37.8 & 87.1 & 84.6 & 71.8 & 25.2 & 45.3 & 75.1 & 64.7 & 80.0 & 49.9 & 53.3 \\
EuroSAT & 37.8 & 87.1 & 84.6 & 71.8 & 97.0 & 45.3 & 75.1 & 64.7 & 80.0 & 49.9 & 53.3 \\
Flowers & 37.8 & 87.1 & 84.6 & 71.8 & 97.0 & 96.3 & 75.1 & 64.7 & 80.0 & 49.9 & 53.8 \\
Food & 37.8 & 87.1 & 84.6 & 71.8 & 97.0 & 96.3 & 89.1 & 64.7 & 80.0 & 49.9 & 53.7 \\
MNIST & 37.8 & 87.1 & 84.6 & 71.8 & 89.2 & 96.3 & 89.1 & 99.1 & 80.0 & 49.9 & 53.7 \\
OxfordPet & 37.8 & 87.1 & 84.6 & 71.8 & 89.2 & 96.3 & 89.1 & 99.1 & 94.5 & 49.9 & 53.8 \\
Cars & 37.8 & 87.1 & 84.6 & 71.8 & 89.2 & 96.3 & 89.1 & 99.1 & 94.5 & 79.9 & 53.4 \\
SUN397 & 37.8 & 87.1 & 84.6 & 71.8 & 89.2 & 96.3 & 89.1 & 99.1 & 94.5 & 79.9 & 76.5 & \colorbox{last}{82.3} \\ \midrule
Avg. & 37.8 & 84.3 & 78.6 & 60.1 & 71.1 & 73.2 & 79.1 & 73.9 & 82.3 & 55.1 & 52.8 & \colorbox{avg}{68.0} \\
\bottomrule
\end{tabular}%
}
\end{table*}

\begin{table*}[t]
\setlength\tabcolsep{5pt}
\centering
\caption{Accuracy (\%) of S-Prompts on the MTIL benchmark with order-I. Each row represents the performance on every dataset of the model trained after the corresponding task. \textcolor{transfer}{Transfer}, \textcolor{avg}{Avg.}, and \textcolor{last}{Last} metrics are shown in color.}
\label{tab:complete_res_sprompts}
\fontsize{8pt}{10pt}\selectfont
\resizebox{\textwidth}{!}{
\begin{tabular}{@{}lcccccccccccc@{}}
\toprule
& \rot{Aircraft} & \rot{Caltech101} & \rot{CIFAR100} & \rot{DTD} & \rot{EuroSAT} & \rot{Flowers} & \rot{Food} & \rot{MNIST} & \rot{OxfordPet} & \rot{Cars} & \rot{SUN397} \\ \midrule
Transfer & & 67.3 & 49.4 & 26.4 & 39.7 & 47.1 & 70.2 & 34.3 & 78.9 & 56.7 & 52.2 & \colorbox{transfer}{52.2} \\ \midrule
Aircraft & 37.5 & 67.3 & 40.1 & 12.8 & 23.5 & 15.3 & 41.1 & 37.5 & 47.7 & 57.9 & 37.9 \\
Caltech101 & 37.5 & 95.0 & 58.8 & 33.2 & 36.5 & 56.5 & 77.3 & 39.1 & 83.4 & 56.5 & 54.5 \\
CIFAR100 & 37.5 & 95.0 & 83.7 & 33.2 & 49.3 & 56.5 & 77.3 & 32.7 & 83.4 & 56.5 & 54.5 \\
DTD & 37.5 & 95.0 & 83.7 & 70.2 & 49.3 & 53.5 & 75.2 & 32.7 & 83.4 & 56.5 & 53.7 \\
EuroSAT & 37.5 & 95.0 & 83.7 & 70.2 & 97.5 & 53.5 & 75.2 & 32.7 & 83.4 & 56.5 & 53.7 \\
Flowers & 37.5 & 95.0 & 83.7 & 70.2 & 97.5 & 96.5 & 75.2 & 32.7 & 83.4 & 56.5 & 53.7 \\
Food & 37.5 & 95.0 & 83.7 & 70.2 & 97.5 & 96.5 & 89.0 & 32.7 & 83.4 & 56.5 & 53.7 \\
MNIST & 37.5 & 95.0 & 83.7 & 70.2 & 97.5 & 96.5 & 89.0 & 99.1 & 83.4 & 56.5 & 53.7 \\
OxfordPet & 37.5 & 95.0 & 83.7 & 70.2 & 97.5 & 96.5 & 89.0 & 99.1 & 94.0 & 56.5 & 53.7 \\
Cars & 37.5 & 95.0 & 83.7 & 70.2 & 97.5 & 96.5 & 89.0 & 99.1 & 94.0 & 79.5 & 53.7 \\
SUN397 & 37.5 & 95.0 & 83.7 & 70.2 & 97.5 & 96.5 & 89.0 & 99.1 & 94.0 & 79.5 & 75.8 & \colorbox{last}{83.4} \\ \midrule
Avg. & 37.5 & 92.5 & 77.5 & 58.2 & 76.4 & 74.1 & 78.8 & 57.9 & 83.0 & 60.8 & 54.4 & \colorbox{avg}{68.3} \\
\bottomrule
\end{tabular}%
}
\end{table*}

\begin{table*}[t]
\setlength\tabcolsep{5pt}
\centering
\caption{Accuracy (\%) of our DIKI on the MTIL benchmark with order-II. Each row represents the performance on every dataset of the model trained after the corresponding task. \textcolor{transfer}{Transfer}, \textcolor{avg}{Avg.}, and \textcolor{last}{Last} metrics are shown in color.}
\label{tab:complete_res_ours_order2}
\fontsize{8pt}{10pt}\selectfont
\resizebox{\textwidth}{!}{
\begin{tabular}{@{}lcccccccccccc@{}}
\toprule
& \rot{Cars} & \rot{Food} & \rot{MNIST} & \rot{OxfordPet} & \rot{Flowers} & \rot{SUN397} & \rot{Aircraft} & \rot{Caltech101} & \rot{DTD} & \rot{EuroSAT} & \rot{CIFAR100} \\ \midrule
Transfer & & 85.8 & 59.8 & 89.1 & 71.8 & 62.6 & 24.3 & 93.3 & 42.7 & 46.8 & 67.8 & \colorbox{transfer}{64.4} \\ \midrule
Cars & 81.9 & 85.8 & 59.7 & 89.2 & 71.5 & 62.6 & 24.9 & 92.9 & 44.0 & 47.6 & 68.4 \\
Food & 81.9 & 89.2 & 59.9 & 89.1 & 71.9 & 62.7 & 24.9 & 93.1 & 43.8 & 47.7 & 68.5 \\
MNIST & 81.9 & 89.2 & 99.3 & 89.1 & 71.9 & 62.7 & 24.9 & 93.1 & 43.8 & 47.7 & 68.5 \\
OxfordPet & 81.9 & 89.2 & 99.3 & 94.2 & 71.9 & 62.6 & 24.9 & 93.0 & 43.8 & 47.7 & 68.4 \\
Flowers & 81.9 & 89.2 & 99.3 & 94.2 & 96.7 & 62.6 & 24.9 & 93.0 & 44.0 & 47.7 & 68.4 \\
SUN397 & 81.9 & 89.2 & 99.3 & 94.2 & 96.8 & 76.7 & 21.2 & 94.0 & 40.8 & 44.6 & 67.4 \\
Aircraft & 81.9 & 89.2 & 99.3 & 94.2 & 96.8 & 76.7 & 46.3 & 94.0 & 40.8 & 44.6 & 67.4 \\
Caltech101 & 81.9 & 89.2 & 99.3 & 94.3 & 96.8 & 76.7 & 46.3 & 95.9 & 40.7 & 44.7 & 67.2 \\
DTD & 81.9 & 89.2 & 99.3 & 94.3 & 96.8 & 76.7 & 46.3 & 95.9 & 74.8 & 48.4 & 66.8 \\
EuroSAT & 81.9 & 89.2 & 99.3 & 94.3 & 96.8 & 76.7 & 46.3 & 95.9 & 74.8 & 98.2 & 66.8 \\
CIFAR100 & 81.9 & 89.2 & 99.4 & 94.3 & 96.8 & 76.7 & 46.3 & 95.9 & 74.8 & 98.3 & 86.6 & \colorbox{last}{85.5} \\ \midrule
Avg. & 81.9 & 88.9 & 92.1 & 92.8 & 87.7 & 70.3 & 34.3 & 94.2 & 51.5 & 56.1 & 69.5 & \colorbox{avg}{74.5} \\
\bottomrule
\end{tabular}%
}
\end{table*}

\begin{table*}[t]
\setlength\tabcolsep{5pt}
\centering
\caption{Accuracy (\%) of L2P on the MTIL benchmark with order-II. Each row represents the performance on every dataset of the model trained after the corresponding task. \textcolor{transfer}{Transfer}, \textcolor{avg}{Avg.}, and \textcolor{last}{Last} metrics are shown in color.}
\label{tab:complete_res_l2p_order2}
\fontsize{8pt}{10pt}\selectfont
\resizebox{\textwidth}{!}{
\begin{tabular}{@{}lcccccccccccc@{}}
\toprule
& \rot{Cars} & \rot{Food} & \rot{MNIST} & \rot{OxfordPet} & \rot{Flowers} & \rot{SUN397} & \rot{Aircraft} & \rot{Caltech101} & \rot{DTD} & \rot{EuroSAT} & \rot{CIFAR100} \\ \midrule
Transfer & & 70.6 & 30.7 & 78.3 & 42.8 & 38.3 & 17.4 & 75.3 & 27.4 & 23.1 & 20.7 & \colorbox{transfer}{42.5} \\ \midrule
Cars & 80.1 & 70.6 & 41.1 & 67.6 & 42.1 & 44.6 & 17.5 & 79.0 & 27.8 & 24.3 & 51.8 \\
Food & 80.1 & 89.1 & 20.3 & 83.7 & 56.9 & 50.1 & 17.5 & 84.7 & 28.9 & 25.1 & 52.0 \\
MNIST & 80.1 & 89.1 & 99.1 & 83.7 & 56.9 & 29.8 & 17.5 & 44.2 & 14.4 & 12.7 & 12.9 \\
OxfordPet & 80.1 & 89.1 & 99.1 & 93.8 & 15.2 & 30.0 & 17.5 & 69.4 & 14.4 & 12.7 & 12.9 \\
Flowers & 80.1 & 89.1 & 99.1 & 93.8 & 96.2 & 37.1 & 17.5 & 77.8 & 27.8 & 12.7 & 12.9 \\
SUN397 & 80.1 & 89.1 & 99.1 & 93.8 & 96.2 & 76.5 & 16.8 & 89.7 & 35.9 & 29.8 & 12.9 \\
Aircraft & 80.1 & 89.1 & 99.1 & 93.8 & 96.2 & 76.5 & 40.1 & 82.3 & 35.9 & 29.8 & 12.9 \\
Caltech101 & 80.1 & 89.1 & 99.1 & 93.8 & 96.2 & 76.5 & 40.1 & 86.9 & 33.8 & 29.8 & 12.9 \\
DTD & 80.1 & 89.1 & 99.1 & 93.8 & 96.2 & 76.5 & 40.1 & 86.9 & 73.5 & 30.8 & 12.9 \\
EuroSAT & 80.1 & 89.1 & 99.1 & 93.8 & 96.2 & 76.5 & 40.1 & 86.9 & 73.5 & 86.3 & 12.9 \\
CIFAR100 & 80.1 & 89.1 & 99.1 & 93.8 & 96.2 & 76.5 & 40.1 & 86.9 & 73.5 & 86.3 & 84.2 & \colorbox{last}{82.3} \\ \midrule
Avg. & 80.1 & 87.4 & 86.7 & 89.6 & 76.8 & 59.1 & 27.7 & 79.5 & 39.9 & 34.6 & 26.5 & \colorbox{avg}{62.5} \\
\bottomrule
\end{tabular}%
}
\end{table*}

\begin{table*}[t]
\setlength\tabcolsep{5pt}
\centering
\caption{Accuracy (\%) of DualPrompt on the MTIL benchmark with order-II. Each row represents the performance on every dataset of the model trained after the corresponding task. \textcolor{transfer}{Transfer}, \textcolor{avg}{Avg.}, and \textcolor{last}{Last} metrics are shown in color.}
\label{tab:complete_res_dualprompt_order2}
\fontsize{8pt}{10pt}\selectfont
\resizebox{\textwidth}{!}{
\begin{tabular}{@{}lcccccccccccc@{}}
\toprule
& \rot{Cars} & \rot{Food} & \rot{MNIST} & \rot{OxfordPet} & \rot{Flowers} & \rot{SUN397} & \rot{Aircraft} & \rot{Caltech101} & \rot{DTD} & \rot{EuroSAT} & \rot{CIFAR100} \\ \midrule
Transfer & & 79.9 & 46.9 & 85.2 & 51.3 & 45.1 & 9.3 & 82.7 & 29.9 & 42.9 & 47.2 & \colorbox{transfer}{52.1} \\ \midrule
Cars & 78.6 & 79.9 & 47.7 & 82.8 & 50.1 & 48.7 & 9.3 & 84.2 & 29.4 & 49.7 & 61.7 \\
Food & 78.6 & 89.3 & 46.2 & 86.5 & 53.4 & 54.5 & 9.3 & 87.6 & 28.7 & 51.5 & 64.2 \\
MNIST & 78.6 & 89.3 & 99.2 & 86.5 & 53.4 & 42.4 & 9.3 & 80.4 & 23.9 & 28.6 & 43.3 \\
OxfordPet & 78.6 & 89.3 & 99.2 & 94.1 & 48.4 & 38.4 & 9.3 & 76.5 & 23.9 & 28.6 & 43.3 \\
Flowers & 78.6 & 89.3 & 99.2 & 94.1 & 96.5 & 41.4 & 9.3 & 76.3 & 26.8 & 28.6 & 43.3 \\
SUN397 & 78.6 & 89.3 & 99.2 & 94.1 & 96.5 & 76.8 & 9.3 & 90.2 & 35.9 & 50.0 & 43.3 \\
Aircraft & 78.6 & 89.3 & 99.2 & 94.1 & 96.5 & 76.8 & 39.8 & 83.8 & 35.9 & 50.0 & 43.3 \\
Caltech101 & 78.6 & 89.3 & 99.2 & 94.1 & 96.5 & 76.8 & 39.8 & 89.0 & 34.5 & 50.0 & 43.3 \\
DTD & 78.6 & 89.3 & 99.2 & 94.1 & 96.5 & 76.8 & 39.8 & 89.0 & 71.6 & 49.4 & 43.3 \\
EuroSAT & 78.6 & 89.3 & 99.2 & 94.1 & 96.5 & 76.8 & 39.8 & 89.0 & 71.6 & 90.7 & 43.3 \\
CIFAR100 & 78.6 & 89.3 & 99.2 & 94.1 & 96.5 & 76.8 & 39.8 & 89.0 & 71.6 & 90.7 & 84.9 & \colorbox{last}{82.8} \\ \midrule
Avg. & 78.6 & 88.4 & 89.7 & 91.7 & 80.0 & 62.4 & 23.2 & 85.0 & 41.3 & 51.6 & 50.7 & \colorbox{avg}{67.5} \\
\bottomrule
\end{tabular}%
}
\end{table*}

\begin{table*}[t]
\setlength\tabcolsep{5pt}
\centering
\caption{Accuracy (\%) of S-Prompts on the MTIL benchmark with order-II. Each row represents the performance on every dataset of the model trained after the corresponding task. \textcolor{transfer}{Transfer}, \textcolor{avg}{Avg.}, and \textcolor{last}{Last} metrics are shown in color.}
\label{tab:complete_res_sprompts_order2}
\fontsize{8pt}{10pt}\selectfont
\resizebox{\textwidth}{!}{
\begin{tabular}{@{}lcccccccccccc@{}}
\toprule
& \rot{Cars} & \rot{Food} & \rot{MNIST} & \rot{OxfordPet} & \rot{Flowers} & \rot{SUN397} & \rot{Aircraft} & \rot{Caltech101} & \rot{DTD} & \rot{EuroSAT} & \rot{CIFAR100} \\ \midrule
Transfer & & 59.8 & 46.2 & 67.7 & 47.5 & 43.8 & 13.5 & 76.8 & 31.4 & 22.6 & 43.5 & \colorbox{transfer}{45.3} \\ \midrule
Cars & 79.2 & 59.8 & 60.1 & 55.0 & 26.9 & 38.0 & 13.4 & 70.3 & 27.5 & 14.3 & 39.7 \\
Food & 79.2 & 89.1 & 32.3 & 74.0 & 56.1 & 47.2 & 13.4 & 76.6 & 27.7 & 18.1 & 53.5 \\
MNIST & 79.2 & 89.1 & 99.1 & 74.0 & 56.1 & 46.8 & 13.4 & 72.6 & 30.5 & 18.7 & 42.7 \\
OxfordPet & 79.2 & 89.1 & 99.1 & 94.3 & 50.9 & 44.3 & 13.4 & 66.2 & 31.4 & 18.7 & 42.7 \\
Flowers & 79.2 & 89.1 & 99.1 & 94.3 & 95.8 & 42.5 & 13.4 & 77.8 & 27.7 & 18.7 & 42.7 \\
SUN397 & 79.2 & 89.1 & 99.1 & 94.3 & 95.8 & 76.3 & 13.9 & 91.3 & 35.5 & 29.4 & 42.7 \\
Aircraft & 79.2 & 89.1 & 99.1 & 94.3 & 95.8 & 76.3 & 39.9 & 83.0 & 35.5 & 29.4 & 42.7 \\
Caltech101 & 79.2 & 89.1 & 99.1 & 94.3 & 95.8 & 76.3 & 39.9 & 95.5 & 35.2 & 29.4 & 42.7 \\
DTD & 79.2 & 89.1 & 99.1 & 94.3 & 95.8 & 76.3 & 39.9 & 95.5 & 70.1 & 27.1 & 42.7 \\
EuroSAT & 79.2 & 89.1 & 99.1 & 94.3 & 95.8 & 76.3 & 39.9 & 95.5 & 70.1 & 97.6 & 42.7 \\
CIFAR100 & 79.2 & 89.1 & 99.1 & 94.3 & 95.8 & 76.3 & 39.9 & 95.5 & 70.1 & 97.6 & 84.4 & \colorbox{last}{83.8} \\ \midrule
Avg. & 79.2 & 86.5 & 89.5 & 87.0 & 78.2 & 61.5 & 25.5 & 83.6 & 41.9 & 36.3 & 47.2 & \colorbox{avg}{65.1} \\
\bottomrule
\end{tabular}%
}
\end{table*}

\begin{table*}[t]
\setlength\tabcolsep{6pt}
\centering
\caption{Accuracy (\%) of our DIKI on the MTIL-FS benchmark with order-I. Each row represents the performance on every dataset of the model trained after the corresponding task. \textcolor{transfer}{Transfer}, \textcolor{avg}{Avg.}, and \textcolor{last}{Last} metrics are shown in color.}
\label{tab:complete_res_ours_fs}
\fontsize{8pt}{10pt}\selectfont
\resizebox{0.8\textwidth}{!}{
\begin{tabular}{@{}lccccccccc@{}}
\toprule
& \rot{Aircraft} & \rot{Caltech101} & \rot{CIFAR100} & \rot{DTD} & \rot{Flowers} & \rot{Food} & \rot{Cars} & \rot{SUN397} \\ \midrule
Transfer & & 92.7 & 68.8 & 44.1 & 70.0 & 86.2 & 65.1 & 65.5 & \colorbox{transfer}{70.3} \\ \midrule
Aircraft & 41.4 & 92.7 & 68.4 & 43.9 & 71.3 & 85.8 & 65.8 & 62.5 \\
Caltech101 & 41.3 & 95.7 & 69.2 & 44.2 & 69.5 & 86.3 & 64.9 & 66.0 \\
CIFAR100 & 41.3 & 95.7 & 79.0 & 44.2 & 69.5 & 86.3 & 64.9 & 66.0 \\
DTD & 41.3 & 95.7 & 79.0 & 67.1 & 69.5 & 86.3 & 64.9 & 66.0 \\
Flowers & 41.3 & 95.7 & 79.0 & 67.1 & 94.5 & 86.3 & 64.9 & 66.0 \\
Food & 41.3 & 95.7 & 79.0 & 67.1 & 94.5 & 86.8 & 64.9 & 66.0 \\
Cars & 41.3 & 95.7 & 79.0 & 67.1 & 94.5 & 86.8 & 77.5 & 66.0 \\
SUN397 & 41.3 & 95.6 & 79.0 & 67.3 & 94.4 & 86.8 & 77.6 & 74.4 & \colorbox{last}{77.1} \\ \midrule
Avg. & 41.3 & 95.3 & 76.5 & 58.5 & 82.2 & 86.4 & 68.2 & 66.6 & \colorbox{avg}{71.9} \\
\bottomrule
\end{tabular}%
}
\end{table*}

\begin{table*}[t]
\setlength\tabcolsep{6pt}
\centering
\caption{Accuracy (\%) of L2P on the MTIL-FS benchmark with order-I. Each row represents the performance on every dataset of the model trained after the corresponding task. \textcolor{transfer}{Transfer}, \textcolor{avg}{Avg.}, and \textcolor{last}{Last} metrics are shown in color.}
\label{tab:complete_res_l2p_fs}
\fontsize{8pt}{10pt}\selectfont
\resizebox{0.8\textwidth}{!}{
\begin{tabular}{@{}lccccccccc@{}}
\toprule
& \rot{Aircraft} & \rot{Caltech101} & \rot{CIFAR100} & \rot{DTD} & \rot{Flowers} & \rot{Food} & \rot{Cars} & \rot{SUN397} \\ \midrule
Transfer & & 66.7 & 54.3 & 30.6 & 47.3 & 71.5 & 54.6 & 52.4 & \colorbox{transfer}{53.9} \\ \midrule
Aircraft & 30.1 & 66.7 & 44.3 & 23.0 & 32.4 & 47.8 & 49.9 & 32.9 \\
Caltech101 & 30.1 & 87.1 & 64.2 & 34.5 & 53.5 & 77.6 & 55.6 & 56.5 \\
CIFAR100 & 30.1 & 87.1 & 75.3 & 34.5 & 53.5 & 77.6 & 55.6 & 56.5 \\
DTD & 30.1 & 87.1 & 75.3 & 64.7 & 49.7 & 77.3 & 55.6 & 55.6 \\
Flowers & 30.1 & 87.1 & 75.3 & 64.7 & 91.9 & 77.3 & 55.6 & 55.0 \\
Food & 30.1 & 87.1 & 75.3 & 64.7 & 91.9 & 86.4 & 55.6 & 55.0 \\
Cars & 30.1 & 87.1 & 75.3 & 64.7 & 91.9 & 86.4 & 76.2 & 55.5 \\
SUN397 & 30.1 & 87.1 & 75.3 & 64.7 & 91.9 & 86.4 & 76.2 & 74.7 & \colorbox{last}{73.3} \\ \midrule
Avg. & 30.2 & 84.5 & 70.1 & 51.9 & 69.6 & 77.1 & 60.0 & 55.2 & \colorbox{avg}{62.3} \\
\bottomrule
\end{tabular}%
}
\end{table*}

\begin{table*}[t]
\setlength\tabcolsep{6pt}
\centering
\caption{Accuracy (\%) of DualPrompt on the MTIL-FS benchmark with order-I. Each row represents the performance on every dataset of the model trained after the corresponding task. \textcolor{transfer}{Transfer}, \textcolor{avg}{Avg.}, and \textcolor{last}{Last} metrics are shown in color.}
\label{tab:complete_res_dualprompt_fs}
\fontsize{8pt}{10pt}\selectfont
\resizebox{0.8\textwidth}{!}{
\begin{tabular}{@{}lccccccccc@{}}
\toprule
& \rot{Aircraft} & \rot{Caltech101} & \rot{CIFAR100} & \rot{DTD} & \rot{Flowers} & \rot{Food} & \rot{Cars} & \rot{SUN397} \\ \midrule
Transfer & & 78.8 & 64.4 & 32.0 & 51.7 & 77.5 & 49.4 & 51.3 & \colorbox{transfer}{57.9} \\ \midrule
Aircraft & 36.5 & 78.8 & 61.5 & 28.4 & 51.6 & 79.4 & 57.5 & 52.2 \\
Caltech101 & 36.5 & 91.0 & 67.4 & 33.8 & 51.5 & 75.0 & 47.8 & 51.5 \\
CIFAR100 & 36.5 & 91.0 & 75.1 & 33.8 & 51.5 & 75.0 & 47.8 & 51.5 \\
DTD & 36.5 & 91.0 & 75.1 & 65.1 & 52.2 & 79.2 & 47.8 & 51.1 \\
Flowers & 36.5 & 91.0 & 75.1 & 65.1 & 92.9 & 79.2 & 47.8 & 51.1 \\
Food & 36.5 & 91.0 & 75.1 & 65.1 & 92.9 & 86.2 & 47.8 & 51.1 \\
Cars & 36.5 & 91.0 & 75.1 & 65.1 & 92.9 & 86.2 & 76.2 & 50.7 \\
SUN397 & 36.5 & 91.0 & 75.1 & 65.1 & 92.9 & 86.2 & 76.2 & 74.2 & \colorbox{last}{74.7} \\ \midrule
Avg. & 36.5 & 89.5 & 72.5 & 52.7 & 72.3 & 80.8 & 56.1 & 54.2 & \colorbox{avg}{64.3} \\
\bottomrule
\end{tabular}%
}
\end{table*}

\begin{table*}[t]
\setlength\tabcolsep{6pt}
\centering
\caption{Accuracy (\%) of S-Prompts on the MTIL-FS benchmark with order-I. Each row represents the performance on every dataset of the model trained after the corresponding task. \textcolor{transfer}{Transfer}, \textcolor{avg}{Avg.}, and \textcolor{last}{Last} metrics are shown in color.}
\label{tab:complete_res_sprompts_fs}
\fontsize{8pt}{10pt}\selectfont
\resizebox{0.8\textwidth}{!}{
\begin{tabular}{@{}lccccccccc@{}}
\toprule
& \rot{Aircraft} & \rot{Caltech101} & \rot{CIFAR100} & \rot{DTD} & \rot{Flowers} & \rot{Food} & \rot{Cars} & \rot{SUN397} \\ \midrule
Transfer & & 70.3 & 52.7 & 31.5 & 54.8 & 74.0 & 55.4 & 50.0 & \colorbox{transfer}{55.5} \\ \midrule
Aircraft & 30.6 & 70.3 & 44.5 & 24.5 & 46.2 & 72.6 & 53.7 & 32.3 \\
Caltech101 & 30.6 & 89.2 & 60.8 & 35.0 & 60.2 & 75.7 & 55.7 & 53.8 \\
CIFAR100 & 30.6 & 89.2 & 75.8 & 35.0 & 60.2 & 75.7 & 55.7 & 53.8 \\
DTD & 30.6 & 89.2 & 75.8 & 63.8 & 52.7 & 72.8 & 55.7 & 52.4 \\
Flowers & 30.6 & 89.2 & 75.8 & 63.8 & 93.9 & 72.8 & 55.7 & 52.4 \\
Food & 30.6 & 89.2 & 75.8 & 63.8 & 93.9 & 86.2 & 55.7 & 52.4 \\
Cars & 30.6 & 89.2 & 75.8 & 63.8 & 93.9 & 86.2 & 76.7 & 52.4 \\
SUN397 & 30.6 & 89.2 & 75.8 & 63.8 & 93.9 & 86.2 & 76.7 & 73.9 & \colorbox{last}{73.8} \\ \midrule
Avg. & 30.6 & 86.8 & 70.0 & 51.7 & 74.3 & 78.5 & 60.7 & 53.0 & \colorbox{avg}{63.2} \\
\bottomrule
\end{tabular}%
}
\end{table*}

\begin{table*}[t]
\setlength\tabcolsep{6pt}
\centering
\caption{Accuracy (\%) of ZSCL on the MTIL-FS benchmark with order-I. Each row represents the performance on every dataset of the model trained after the corresponding task. \textcolor{transfer}{Transfer}, \textcolor{avg}{Avg.}, and \textcolor{last}{Last} metrics are shown in color.}
\label{tab:complete_res_zscl_fs}
\fontsize{8pt}{10pt}\selectfont
\resizebox{0.8\textwidth}{!}{
\begin{tabular}{@{}lccccccccc@{}}
\toprule
& \rot{Aircraft} & \rot{Caltech101} & \rot{CIFAR100} & \rot{DTD} & \rot{Flowers} & \rot{Food} & \rot{Cars} & \rot{SUN397} \\ \midrule
Transfer & & 87.3 & 67.7 & 45.4 & 67.8 & 86.6 & 59.7 & 63.4 & \colorbox{transfer}{68.3} \\ \midrule
Aircraft & 41.0 & 87.3 & 67.8 & 45.4 & 68.6 & 88.5 & 63.2 & 64.1 \\
Caltech101 & 38.5 & 91.5 & 67.7 & 45.0 & 65.4 & 85.9 & 59.6 & 62.9 \\
CIFAR100 & 37.1 & 91.4 & 79.5 & 45.7 & 68.6 & 87.3 & 60.0 & 64.7 \\
DTD & 36.0 & 91.2 & 78.6 & 68.6 & 68.5 & 86.4 & 59.3 & 62.9 \\
Flowers & 32.1 & 91.1 & 77.3 & 67.5 & 93.8 & 85.1 & 58.3 & 63.1 \\
Food & 30.0 & 90.9 & 76.8 & 66.5 & 91.7 & 90.0 & 57.7 & 62.8 \\
Cars & 25.7 & 90.2 & 75.4 & 64.6 & 90.8 & 89.2 & 80.1 & 63.3 \\
SUN397 & 27.7 & 90.9 & 74.4 & 64.7 & 90.2 & 89.2 & 80.6 & 74.6 & \colorbox{last}{74.0} \\ \midrule
Avg. & 33.5 & 90.5 & 74.7 & 58.5 & 79.7 & 87.7 & 64.8 & 64.8 & \colorbox{avg}{69.3} \\
\bottomrule
\end{tabular}%
}
\end{table*}

\clearpage

%
%
\bibliographystyle{splncs04}
\bibliography{main}